%% file: 0-main.tex
\documentclass[letterpaper]{article} % DO NOT CHANGE THIS
\usepackage{aaai2026}  % DO NOT CHANGE THIS
\usepackage{times}  % DO NOT CHANGE THIS
\usepackage{helvet}  % DO NOT CHANGE THIS
\usepackage{courier}  % DO NOT CHANGE THIS
\usepackage[hyphens]{url}  % DO NOT CHANGE THIS
\usepackage{graphicx} % DO NOT CHANGE THIS
\usepackage{natbib}  % DO NOT CHANGE THIS AND DO NOT ADD ANY OPTIONS TO IT
\usepackage{caption} % DO NOT CHANGE THIS AND DO NOT ADD ANY OPTIONS TO IT
\usepackage{algorithm}
\usepackage{algorithmic}

\usepackage{amsmath,amsthm,amssymb}
\usepackage{booktabs}
\usepackage{multirow}

\usepackage{xcolor}
\usepackage{colortbl}

\usepackage{newfloat}
\usepackage{listings}
\DeclareCaptionStyle{ruled}{labelfont=normalfont,labelsep=colon,strut=off} % DO NOT CHANGE THIS
\floatstyle{ruled}
\newfloat{listing}{tb}{lst}{}
\floatname{listing}{Listing}
\title{Modality-Balanced Collaborative Distillation for Multi-Modal Domain Generalization}

\author{
    Xiaohan Wang\textsuperscript{\rm 1,2}, Zhangtao Cheng\textsuperscript{\rm 1}, Ting Zhong\textsuperscript{\rm 1}\thanks{Corresponding Authors.}, Leiting Chen\textsuperscript{\rm 1,2}, Fan Zhou\textsuperscript{\rm 1,2}\\
}
\affiliations{
    \textsuperscript{\rm 1}University of Electronic Science and Technology of China \\
    \textsuperscript{\rm 2}Key Laboratory of Intelligent Digital Media Technology of Sichuan Province
    \{xiaohanwang, zhangtao.cheng\}@std.uestc.edu.cn, \{zhongting, richardchen, fan.zhou\}@uestc.edu.cn
}

\usepackage{bibentry}
\newcommand\M{MBCD}

\begin{document}

\maketitle

\begin{abstract}
Weight Averaging (WA) has emerged as a powerful technique for enhancing generalization by promoting convergence to a flat loss landscape, which correlates with stronger out-of-distribution performance. However, applying WA directly to multi-modal domain generalization (MMDG) is challenging: differences in optimization speed across modalities lead WA to overfit to faster-converging ones in early stages, suppressing the contribution of slower yet complementary modalities, thereby hindering effective modality fusion and skewing the loss surface toward sharper, less generalizable minima.
To address this issue, we propose MBCD, a unified collaborative distillation framework that retains WA's flatness-inducing advantages while overcoming its shortcomings in multi-modal contexts. MBCD begins with adaptive modality dropout in the student model to curb early-stage bias toward dominant modalities. A gradient consistency constraint then aligns learning signals between uni-modal branches and the fused representation, encouraging coordinated and smoother optimization. Finally, a WA-based teacher conducts cross-modal distillation by transferring fused knowledge to each uni-modal branch, which strengthens cross-modal interactions and steer convergence toward flatter solutions. 
Extensive experiments on MMDG benchmarks show that MBCD consistently outperforms existing methods, achieving superior accuracy and robustness across diverse unseen domains.
% Code is released at {\color{blue}\url{https://github.com/xiaohanwang01/MBCD}}.
\end{abstract}

\begin{links}
\link{Code}{https://github.com/xiaohanwang01/MBCD}
\end{links}

\section{Introduction}
\input{1-introduction}

\section{Related Work}
\input{2-related_work}

\section{Methodology}
\input{3-methodology}

\section{Experiments}
\input{4-experiment}

\section{Conclusion}

\input{5-conclusion}

\section*{Acknowledgments}
\input{6-acknowledgment}

% \section*{Acknowledgements}
% This work was supported by National Natural Science Foundation of China (Grant No. 62572097, No. 62176043, and No. U22A2097).

% \bigskip
% \noindent Thank you for reading these instructions carefully. We look forward to receiving your electronic files!

\bibliography{aaai2026}

% \newpage
% \section{Reproducibility Checklist}
% \label{Checklist}
% \input{8-checklist}

% \newpage
\appendix
\input{7-appendix}

\end{document}

%% file: 1-introduction.tex
Domain Generalization (DG) \cite{wang2022generalizing} aims to build robust machine learning models that can generalize reliably on unseen target domains, even when trained exclusively on data from a limited set of source domains. In practice, conventional supervised learning models often suffer severe performance degradation when deployed in real-world scenarios, such as autonomous driving \cite{sanchez2023domain,yang2024generalized}, medical image diagnosis \cite{yadav2019deep,yoon2024domain}, action recognition \cite{gong2023mmg}, and industrial fault diagnosis \cite{ragab2022conditional,chen2025applications}. To address these challenges, researchers employ a variety of techniques, including domain alignment \cite{li2018domain,li2025seeking}, data augmentation \cite{volpi2018generalizing,xu2025adversarial}, loss landscape flattening \cite{foret2020sharpness,wu2024cr} and weight averaging (WA)~\cite{izmailov2018averaging,rame2023model}.  With the growing prevalence of multiple modalities in modern AI systems, such as vision-language pairs \cite{radford2021learning} and audio-visual \cite{cheng2024retrieval} streams, multi-modal domain generalization (MMDG) has become an increasingly important yet complex challenge. In MMDG, models must learn to align and fuse heterogeneous modalities in the presence of distribution shifts, which significantly amplifies the complexity of the generalization, especially as models often rely on redundancy \cite{tairedundancy}.
 \begin{figure}[t]
    \centering
    \includegraphics[width=\linewidth]{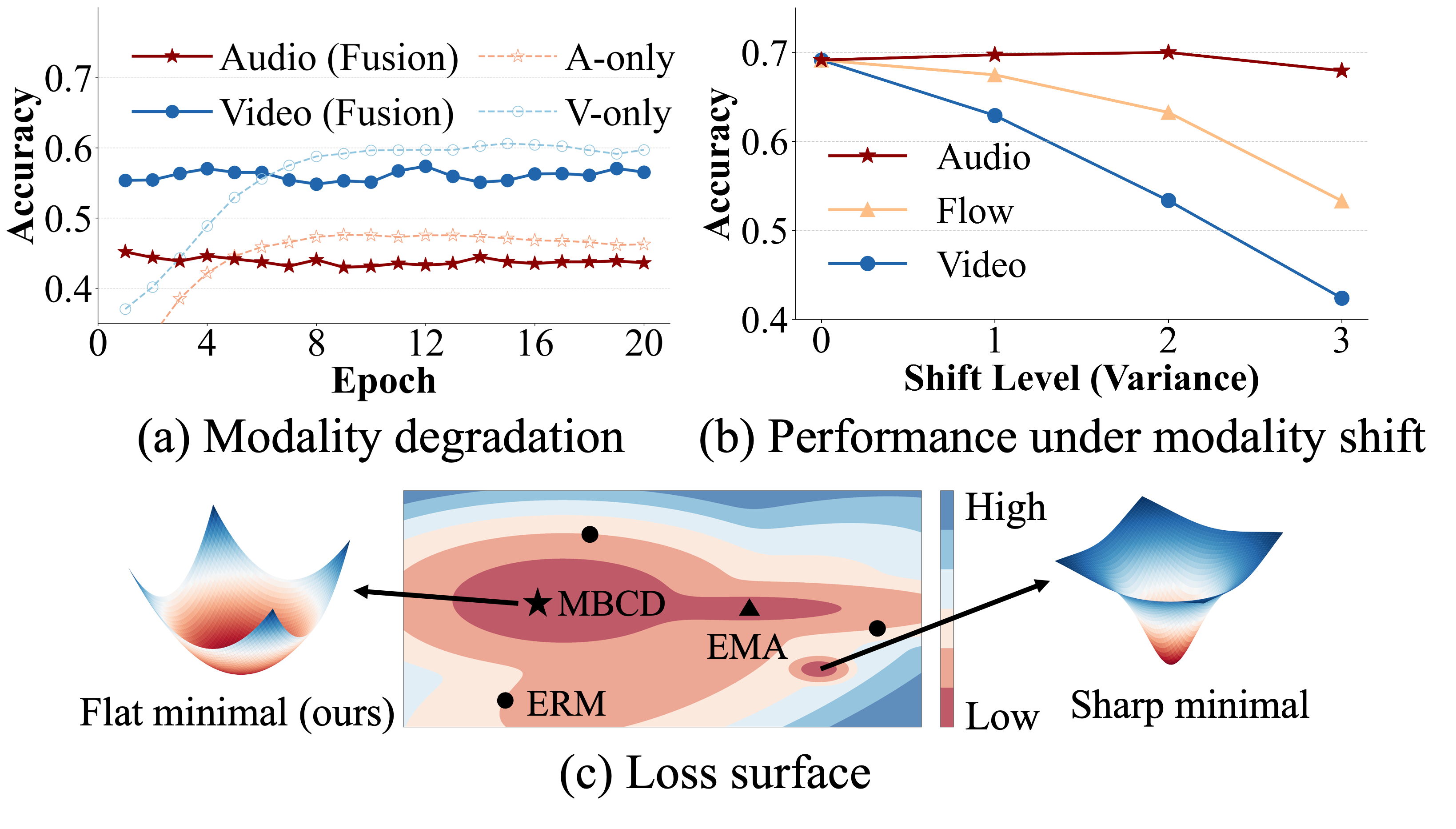}
    \caption{Illustration of EMA's limitations in MMDG. (a) Comparison on the EPIC-Kitchens test set: the uni-modal models are trained from scratch, while the multi-modal model uses 
    post-hoc classifier training.
    % trained post-hoc using a classifier.
    (b) Performance under varying modality shifts induced by Gaussian noise with different variance (shift level); performance drops sharply as the dominant modality (video) is perturbed.
    % shifts. 
    (c) Loss landscape visualization: EMA converges to sharp, biased minima under distributional shifts, while our \M~yields flatter minima, improved robustness.}
    \label{fig:intro}
\end{figure}

Recent work \cite{cha2021swad} has highlighted the importance of loss landscape geometry 
% in enhancing 
for
generalization, particularly focusing on flat minima that 
% correlate with 
offer better 
% robustness under 
resilience to
distribution 
shifts. Among them, Exponential Moving Average (EMA) \cite{li2024switch} 
% has emerged as a simple 
stands out as a straightforward 
and powerful method: by averaging model parameters over training iterations, it smooths the optimization path and encourages convergence to flatter regions, while boosting out-of-distribution performance in uni-modal DG tasks.

% By averaging model parameters over training iterations, EMA implicitly smooths the optimization trajectory, promoting convergence to flatter areas of the loss surface. This property has been shown to significantly improve out-of-distribution performance in uni-modal DG tasks.

% While EMA has proven effective in uni-modal DG, 

However, directly adapting EMA to multi-modal settings introduces new challenges.
% extending it to multi-modal settings introduces new challenges.
The main issue stems from optimization imbalance:
% optimization disparities: 
% different 
modalities 
% often 
converge at different rates due to varying signal strengths and architectural.
% complexity. 
In multi-modal fusion networks, this causes EMA to disproportionately favor fast-converging (and often dominant) modalities during early training. 
Consequently, the contributions of slower yet complementary modalities are suppressed, hindering effective cross-modal fusion and weakening the representational capacity of the model. As shown in Figure~\ref{fig:intro}(a), the performance of the jointly trained model can even lag behind its uni-modal counterparts. This imbalance in optimization dynamics skews the averaged trajectory toward sharper, modality-biased minima (Figure~\ref{fig:intro}(c)), thereby impairing generalization. The vulnerability becomes more pronounced under dominant modality shifts, resulting in significant performance degradation, as illustrated in Figure~\ref{fig:intro}(b). Notably, the accuracy of the weaker modality like audio even increases under mild shifts, highlighting its limited contribution in the fused representation.

% to the fused representation during training.

To bridge this gap, we present \textbf{M}odality-\textbf{B}alanced \textbf{C}ollaborative \textbf{D}istillation (\M), a unified framework that retains 
% the flatness-inducing benefits of EMA 
EMA's flatness benefits
while explicitly mitigating its pitfalls in multi-modal learning. 
% First, we 
MBCD
incorporate adaptive modality dropout into the student model
% , which 
to
dynamically suppresses dominant modalities during training and reduce early-stage over-reliance. 
% Second, we 
It then
introduce a gradient consistency constraint to align the learning dynamics between uni-modal branches and the fused representation, 
% promoting 
enabling 
smoother, more coordinated optimization. Finally, 
% the 
an
EMA-based teacher performs cross-modal distillation,
% transferring 
injecting
fused knowledge into each modality-specific branch, thereby enhancing cross-modal synergy and guiding the model toward flatter and more generalizable solutions. Overall, our contributions can be summarized as follows:
\begin{itemize}
    \item A detailed analysis of EMA's core limitations in MMDG: optimization disparities lead to overfitting on dominant modalities, suppressing cross-modal interactions, and favoring sharper minima that harms generalization.

    % We analyze the core limitation of applying EMA to MMDG. Specifically, we show that optimization disparities across modalities cause EMA to overfit to dominant, fast-converging modalities, suppressing cross-modal interaction and steering the optimization toward sharper minima, ultimately degrading generalization.
    
    \item \M, a unified framework that maintains EMA's strengths while alleviating modality imbalance through collaborative distillation, promoting balanced optimizations and flatter generalizations.

    % We propose \M, a unified framework that preserves the flatness-inducing benefits of EMA while explicitly mitigating modality imbalance. By leveraging collaborative distillation, \M~promotes balanced optimization across modalities and guides the model toward flatter, more generalizable solutions.

    \item  Comprehensive experiments on two public MMDG benchmarks demonstrate the effectiveness of our  proposed \M~across diverse modality combinations, yielding flatter minima and improved generalization.

    % Comprehensive experiments on two public MMDG benchmarks, EPIC-Kitchens and HAC, validate the effectiveness of our \M~framework across diverse modality combinations, yielding flatter minima and improved generalization.
    
\end{itemize}

%% file: 2-related_work.tex
\noindent\textbf{Domain Generalization.} 
This task aims to train models on one or multiple source domains that can generalize to previously unseen target domains \cite{zhou2022domain}. Specifically, domain alignment methods \cite{li2018domain,qu2023modality} focus on learning domain-invariant representations. Moreover, data augmentation methods \cite{zhang2017mixup,qiao2020learning,zheng2024advst,cho2025peer} aim to simulate potential domain shifts by introducing variations at the input or feature level.
Another line of research focuses on flattening the loss landscape \cite{wang2023sharpness,wu2024cr,deng2025asymptotic}, encouraging convergence to flat minima. More recently, weight averaging methods \cite{cha2021swad,arpit2022ensemble,rame2022diverse,rame2023model,javed2024qt}\textemdash such as stochastic weight averaging (SWA) \cite{izmailov2018averaging} and exponential moving average (EMA) \cite{morales2024exponential,ajroldi2025and}\textemdash have emerged as effective techniques for improving DG performance. 

\noindent\textbf{Multi-Modal Domain Generalization.}
While uni-modal DG has been extensively studied, its multi-modal counterpart remains relatively underexplored. Existing efforts remain limited but provide promising directions. \cite{planamente2022domain} align audio and visual features through a relative norm alignment loss to enhance cross-domain robustness and generalizable feature learning. \cite{dong2023simmmdg} disentangles modality-specific and shared features with tailored constraints. \cite{dong2024towards} leverages self-supervised objectives like masked cross-modal translation and multimodal jigsaw puzzles to enhance MMDG. \cite{fan2024cross} promotes consistent flat minima and cross-modal knowledge transfer to enhance modality-specific learning. \cite{huang2025bridging} proposes to mitigate modality asynchrony by learning a highly aligned and unified multi-modal representation space.
However, prior approaches treat all modalities equally, resulting in weaker modalities being underoptimized and convergence to sharper minima. In contrast, our method suppresses dominant modalities, promoting wider minima and enhancing MMDG.
% Despite these advances, the effectiveness of WA techniques in multi-modal learning remains largely unexplored. We bridge this gap by demonstrating that a naive application of EMA can impede effective cross-modal fusion, leading to sharper minima with poor generalization. Our work identifies this fundamental limitation, laying the groundwork for a more balanced and robust optimization strategy in MMDG.

\noindent\textbf{Multi-Modal Imbalance Learning.}
% Multi-modal learning often suffers from modality bias, where dominant modalities suppress weaker ones \cite{wang2020makes,peng2022balanced}. To address this issue, a range of approaches \cite{peng2022balanced,fan2023pmr,zhang2024multimodal,ma2025improving} have been proposed to balance the optimization dynamics.
A range of approaches \cite{peng2022balanced,fan2023pmr,ma2025improving} have been proposed to balance multi-modal learning \cite{wang2020makes}. 
For instance, \cite{zhang2024multimodal} adopts alternating uni-modal optimization to reduce inter-modal interference, while \cite{wei2024enhancing} introduces sample-level modality valuation to encourage cooperation. \cite{huang2025adaptive} slows down learning from dominant modalities during critical phases to mitigate early overfitting.
Other approaches enhance modality representations via auxiliary networks. \cite{du2023uni} distills knowledge from uni-modal experts into multi-modal models. Gradient-based strategies \cite{wei2024mmpareto, guo2024classifier} balance learning by aligning gradient magnitudes or directions, though the former incurs high computation and the latter may suffer from noise.
In contrast, our gradient consistency strategy addresses the mismatch between uni- and multi-modal objectives without explicit gradient manipulation.
% More efforts, such as \cite{wei2024mmpareto} and \cite{guo2024classifier}, attempt to balance multi-modal learning by jointly considering the magnitude and direction of gradients. However, \cite{wei2024mmpareto} requires computationally expensive Pareto front calculations, while \cite{guo2024classifier} approximates gradient similarity via parameter similarity, which may yield misleading results. Compared with them, our uni-modal objective-guided strategy empirically and theoretically alleviates the mismatch between multi-modal and uni-modal optimization objectives without the need for explicit gradient manipulation.

%% file: 3-methodology.tex
\begin{figure*}[t]
    \centering
    \includegraphics[width=\textwidth]{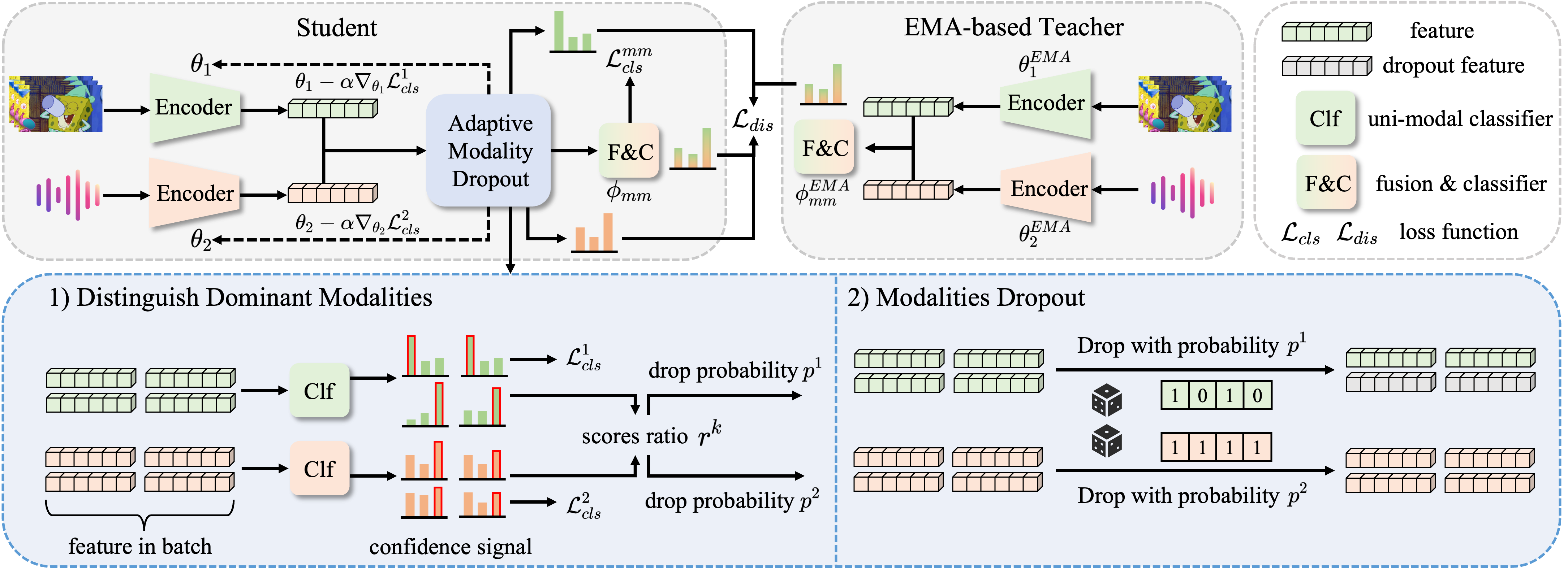}
    \caption{Overall framework of our \M. Our model first performs a uni-modal objective-guided inner-loop update to enhance modality-specific encoders. The updated modality representations are then fused via adaptive modality dropout to mitigate modality imbalance. An EMA-based teacher further guides both uni-modal and fused predictions, promoting stable and modality-balanced learning.}
    \label{fig:model}
\end{figure*}

\subsection{Problem Definition}
We follow the standard formulation of MMDG as described in \cite{dong2023simmmdg}. Let $\mathcal{S}=\{\mathcal{D}_1,\mathcal{D}_2,\cdots,\mathcal{D}_{|S|}\}$ denote a set of source domains used for training, where each domain $\mathcal{D}_j=\{(x_i^{(j)},y_i^{(j)})\}_{i=1}^{N_j}$ contains $N_j$ labeled instances. Each input sample $x_i^{(j)} = \{(x_i^{(j)})_k\}_{k=1}^{M}$ consists of $M$ modalities, and $y_i^{(j)}$ is the corresponding ground-truth label. We assume that the joint distributions differ across domains, i.e., $P_{XY}^i \neq P_{XY}^j, i \neq j$.
The objective of MMDG is to learn a predictive function $f$ that performs well on previously unseen target domains $\mathcal{T}=\{\mathcal{D}_1^{\prime},\mathcal{D}_2^{\prime},\cdots,\mathcal{D}_{|T|}^{\prime}\}$, without accessing any data from $\mathcal{T}$ during training. The optimization objective is formulated as:
\begin{align}
    \min_f\mathbb{E}_{(x,y) \sim P_{\mathcal{T}}}[\mathcal{L}(f(x), y)],
\end{align}
where $\mathbb{E}$ denotes the expectation, and $\mathcal{L}(\cdot,\cdot)$ is a task-specific loss function (e.g., cross-entropy for classification).
In our framework, the parameters of the $M$ modality-specific encoders are denoted by $\theta = \{\theta_1, \ldots, \theta_M\}$, where $\theta_k$ is for the $k$-th modality. The prediction head corresponding to the $k$-th modality is parameterized by $\phi_k$, while the fused multi-modal prediction head is denoted by $\phi_{mm}$. For simplicity, we use $\phi$ to denote $\{\phi_1, \phi_2, \cdots, \phi_M, \phi_{mm}\}$.

% \section{Methodology}

\subsection{\M}

We propose \M, as illustrated in Figure~\ref{fig:model}, which integrates three key components to tackle MMDG:
(1) Adaptive Modality Dropout, which dynamically suppresses dominant modalities to reduce early-stage over-reliance and encourage balanced training;
(2) Gradient Consistency Constraint, which harmonizes uni-modal and multi-modal objectives by aligning their optimization directions, improving training stability and fusion quality;
(3) Collaborative Distillation, which establishes a bidirectional interaction between a student and an EMA-updated teacher. The student learns from the teacher via cross-modal supervision, while simultaneously refining the teacher through online updates, enabling mutual enhancement of generalizable fusion representations.
By integrating these components, \M~consistently achieves superior performance across unseen multi-modal domains.

\subsection{Adaptive Modality Dropout}
Dominant modalities tend to learn faster and may suppress weaker ones during joint training, which can hinder balanced optimization. To mitigate this effect, we propose to adaptively modulate the learning pace of each modality according to its confidence signal. We quantify the confidence of each modality on a mini-batch using the following metric:
\begin{align}
s^k = \sum_{i \in B}\max(\sigma(f^k(x_i^k))),
\end{align}
where $\sigma(\cdot)$ denotes the softmax function, and $f^k(\cdot)$ is the output of the $k$-th modality-specific network, including both the encoder and classifier. The score $s^k$ reflects the overall prediction confidence of modality $k$ on the current mini-batch $B$. Note that this metric does not consider prediction correctness, but rather captures how confident the model is in its outputs, which serves as an indirect signal of its current dominance in learning.

We then quantify the relative learning speed of each modality by comparing its confidence score with those of other modalities:
\begin{align}
    r^k = \frac{1}{M-1} \sum_{j \in \{1, 2, \cdots, M\},j \neq k} \frac{s^k}{s^j}.
\end{align}
A value of $r^k > 1$ indicates that the $k$-th modality is learning faster than average and may dominate the optimization, potentially suppressing under-fitting modalities. To mitigate this imbalance, we introduce an adaptive modality dropout mechanism that selectively suppresses over-dominant modalities. Specifically, we define the dropout mask as:
\begin{align}
    \mathbf{D}^k = 1 - \mathbf{M}^k \sim \mathrm{Bernoulli}(\tanh(\max(r^k - 1, 0))),
\end{align}
where $\mathbf{M}^k$ is a Bernoulli-distributed random variable with success probability $p = \tanh(\max(r^k - 1, 0))$, and $\mathbf{D}^k$ denotes the dropout mask applied to modality $k$. This design probabilistically drops dominant modalities with higher $r^k$ values, effectively slowing down their contribution to the learning process.

\subsection{Gradient Consistency Constraint}
Recent multi-modal learning frameworks \cite{wei2024mmpareto, guo2024classifier} introduce additional uni-modal neural components to enhance the representation capacity of each modality via auxiliary uni-modal tasks. However, naively applying such designs may overlook the gradient conflict between uni-modal and multi-modal learning objectives.
To mitigate this issue, we propose a uni-modal objective-guided learning strategy:
\begin{align}
    \min_{\theta,\phi} \sum_{k=1}^M\mathcal{L}_{cls}&(\theta_k,\phi_k;x_k)
    + \mathcal{L}_{cls}(\bigcup_{k=1}^{M}\theta_k',\phi_{mm};x), \nonumber \\
    & \text{where} \enspace \theta_k' = \theta_k-\alpha\nabla_{\theta_k}\mathcal{L}_{cls}(\theta_k,\phi_k;x_k).
    \label{eq:gradient_consistency}
\end{align}
Here, $\bigcup_{k=1}^{M}\theta_k'$ denotes parameter aggregation, $\mathcal{L}_{cls}(\theta_k;x_k)$ is the uni-modal classification loss for the $k$-th modality, and $\mathcal{L}_{cls}(\bigcup_{k=1}^{M}\theta_k';x)$ corresponds to the fused multi-modal classification loss (implemented via concatenation-based fusion in our framework). Note that $\theta_k'$ is obtained by performing a gradient descent step on the uni-modal loss.

To obtain a more tractable and interpretable objective, we perform a first-order Taylor expansion on the second term of Eq.~\eqref{eq:gradient_consistency} around $\theta_k$. This derivation
% , detailed in the Appendix,
yields a new and equivalent learning objective:
\begin{align}
    \min_{\theta,\phi}&\sum_{k=1}^M\mathcal{L}_{cls}(\theta_k,\phi_k;x_k) + \mathcal{L}_{cls}(\theta,\phi_{mm};x) - \nonumber \\ 
    &\sum_{k=1}^{M}\alpha\nabla_{\theta_k}\mathcal{L}_{cls}(\theta_k,\phi_k;x_k) \cdot \nabla_{\theta_k}\mathcal{L}_{cls}(\theta,\phi;x).
\end{align}

This reformulated objective provides a clear interpretation of our strategy. It aims to minimize both the uni-modal and multi-modal losses while simultaneously maximizing the inner product between their respective gradients. Maximizing this inner product encourages the gradients to be aligned, thereby promoting consistency in the update directions of the uni-modal and multi-modal objectives. Essentially, this approach can be viewed as a form of gradient matching, which fosters coordination between the different learning tasks and effectively addresses the initial problem of gradient conflict.

Additionally, raw features from disparate encoders often exhibit misaligned numerical ranges and variances, which can destabilize joint optimization, suppress weaker modalities, or bias fusion toward dominant modalities. Inspired by the success of Layer Normalization (LayerNorm) \cite{xu2019understanding} in stabilizing single-modality deep networks (e.g., Transformers), we add LayerNorm to each modality’s encoded features.

\subsection{Collaborative Distillation}
EMA is a simple yet effective technique that improves model generalization by smoothing parameter updates over the training trajectory. It has been widely used in various training paradigms to stabilize optimization, suppress noise in parameter updates, and implicitly form a temporal ensemble of models. Formally, the EMA parameters at training step $t$ are computed as:
\begin{align}
    \Theta_t^{EMA} = \beta\Theta_{t-1}^{EMA} + (1-\beta)\Theta_t,
    \label{eq:ema}
\end{align}
where $\Theta_t$ denotes the model parameters at step $t$ updated by the optimizer, $\Theta_t^{EMA}$ represents the exponentially averaged parameters, and $\beta \in [0, 1)$ is a smoothing factor that controls the decay rate of historical information. A larger $\beta$ places more emphasis on past parameter states, resulting in smoother updates. In our framework, EMA is applied to both modality-specific encoder parameters $\theta^k$ and the fused prediction head parameter $\phi^{mm}$. This strategy forms an implicit ensemble of the model over time, which improves robustness to domain shifts.

We then propose the collaborative distillation framework that simultaneously enhances generalization and promotes cross-modal fusion through a bidirectional interaction between the student and an EMA-based teacher model. Our distillation targets two key objectives: (1) enhancing generalization via an EMA-based teacher, and (2) promoting cross-modal interaction by distilling fused knowledge into uni-modal branches. The distillation loss is defined as:
\begin{align}
\mathcal{L}_{dis} = D_{\text{KL}}(p^{EMA}||p^{mm}) + \sum_{m=1}^M D_{\text{KL}}(p^{EMA}||p^m),
\end{align}
where $p^{EMA}$  is the teacher’s fused prediction, $p^{mm}$ is the student’s multi-modal output, and $p^m$ denotes predictions from each uni-modal branch. The KL divergence terms measure the discrepancy between student predictions and the teacher’s fused output. This formulation dynamically monitors modality-wise consistency and mitigates imbalance by penalizing divergence from the teacher.

With cross-modal distillation, our framework forms a collaborative learning loop: the teacher offers stable cross-modal supervision, while the student, through its improved predictions, continuously refines the teacher via online updates. This bidirectional interaction enables mutual enhancement of generalizable fusion representations.

\subsection{Final Loss}
The final loss is obtained as the weighted sum of the previously defined losses:
\begin{align}
    \mathcal{L} = \mathcal{L}_{cls}(\theta^{\prime},\phi_{mm};x) + \sum_{k=1}^M\mathcal{L}(\theta_k,\phi_k;x_k) + \lambda\mathcal{L}_{dis},
\end{align}
where $\lambda$ is the coefficient of distillation loss $\mathcal{L}_{dis}$.

%% file: 4-experiment.tex
\begin{table*}[t]
    \centering
    \resizebox{\textwidth}{!}{
    \begin{tabular}{lccccccccccc}
        \toprule
        \multirow{2.5}{*}{\textbf{Method}} & \multicolumn{3}{c}{\textbf{Modality}} & \multicolumn{4}{c}{\textbf{EPIC-Kitchens}} & \multicolumn{4}{c}{\textbf{HAC}} \\
        \cmidrule(r){2-4} \cmidrule(r){5-8} \cmidrule(r){9-12} \multicolumn{1}{c}{} & \multicolumn{1}{c}{Video} & \multicolumn{1}{c}{Audio} & \multicolumn{1}{c}{Flow} & \multicolumn{1}{c}{D2,D3$\rightarrow$D1} & \multicolumn{1}{c}{D1,D3$\rightarrow$D2} & \multicolumn{1}{c}{D1,D2$\rightarrow$D3} & \multicolumn{1}{c}{Avg.} & \multicolumn{1}{c}{A,C$\rightarrow$H} & \multicolumn{1}{c}{H,C$\rightarrow$A} & \multicolumn{1}{c}{H,A$\rightarrow$C} & \multicolumn{1}{c}{Avg.} \\
        \midrule
        ERM & $\checkmark$ & $\checkmark$ & 
        & 52.04\scriptsize{$\pm$0.60} & 60.73\scriptsize{$\pm$1.69} & 57.92\scriptsize{$\pm$1.40} & 56.90
        & 72.96\scriptsize{$\pm$0.69} & 74.10\scriptsize{$\pm$3.65} & 51.96\scriptsize{$\pm$1.73} & 66.34  \\
        RNA-Net & $\checkmark$ & $\checkmark$ & 
        & 52.21\scriptsize{$\pm$1.20} & 60.06\scriptsize{$\pm$2.06} & 56.17\scriptsize{$\pm$1.87} & 56.15
        & 73.42\scriptsize{$\pm$1.16} & 73.33\scriptsize{$\pm$1.16} & 50.06\scriptsize{$\pm$1.16} & 65.60  \\
        SimMMDG & $\checkmark$ & $\checkmark$ & 
        & 56.35\scriptsize{$\pm$2.19} & 64.52\scriptsize{$\pm$0.72} & 58.21\scriptsize{$\pm$1.94} & 59.69
        & 76.54\scriptsize{$\pm$2.40} & 75.79\scriptsize{$\pm$2.14} & 48.71\scriptsize{$\pm$1.61} & 67.01 \\
        MOOSA & $\checkmark$ & $\checkmark$ & 
        & 54.64\scriptsize{$\pm$3.73} & 64.93\scriptsize{$\pm$1.36} & 61.60\scriptsize{$\pm$0.51} & 60.39
        & 74.29\scriptsize{$\pm$1.40} & 75.90\scriptsize{$\pm$0.55} & 52.79\scriptsize{$\pm$2.49} & 67.66 \\
        CMRF & $\checkmark$ & $\checkmark$ & 
        & 56.77\scriptsize{$\pm$0.30} & 65.21\scriptsize{$\pm$0.31} & 61.79\scriptsize{$\pm$0.33} & 61.26
        & 79.21\scriptsize{$\pm$0.69} & \textbf{79.07\scriptsize{$\pm$0.82}} & \textbf{55.48\scriptsize{$\pm$1.41}} & \textbf{71.25} \\
        
        \textbf{\M} & $\checkmark$ & $\checkmark$ & &
        \textbf{58.06\scriptsize{$\pm$0.50}} & \textbf{68.10\scriptsize{$\pm$0.35}} & \textbf{63.31\scriptsize{$\pm$0.21}} & \textbf{63.16} & 
        \textbf{80.08\scriptsize{$\pm$0.44}} & 78.81\scriptsize{$\pm$0.72} & 53.09\scriptsize{$\pm$1.84} & 70.66 \\
        \midrule

        ERM & $\checkmark$ &  & $\checkmark$ 
        & 56.88\scriptsize{$\pm$0.26} & 65.74\scriptsize{$\pm$1.42} & 58.02\scriptsize{$\pm$1.32} & 60.21
        & 74.89\scriptsize{$\pm$1.12} & 72.33\scriptsize{$\pm$1.51} & 43.35\scriptsize{$\pm$6.35} & 63.52  \\
        RNA-Net & $\checkmark$ &  & $\checkmark$  
        & 59.29\scriptsize{$\pm$0.54} & 66.37\scriptsize{$\pm$1.30} & 57.85\scriptsize{$\pm$2.02} & 61.17
        & 77.19\scriptsize{$\pm$4.33} & 74.58\scriptsize{$\pm$4.33} & 43.11\scriptsize{$\pm$4.33} & 64.96 \\
        SimMMDG & $\checkmark$ &  & $\checkmark$ 
        & 57.99\scriptsize{$\pm$2.29} & 67.59\scriptsize{$\pm$0.42} & 55.85\scriptsize{$\pm$1.88} & 60.48
        & 77.48\scriptsize{$\pm$1.47} & 73.99\scriptsize{$\pm$1.76} & 51.78\scriptsize{$\pm$6.05} & 67.75   \\
        MOOSA & $\checkmark$ & & $\checkmark$ 
        & 59.16\scriptsize{$\pm$1.36} & 66.44\scriptsize{$\pm$0.98} & 61.43\scriptsize{$\pm$1.14} & 62.34
        & 75.25\scriptsize{$\pm$2.25} & 75.68\scriptsize{$\pm$0.51} & 51.75\scriptsize{$\pm$2.66} & 67.56 \\
        CMRF & $\checkmark$ &  & $\checkmark$ 
        & 63.41\scriptsize{$\pm$0.35} & 69.48\scriptsize{$\pm$0.80} & 61.52\scriptsize{$\pm$0.53} & 64.80
        & 80.32\scriptsize{$\pm$0.68} & 77.08\scriptsize{$\pm$1.52} & 51.87\scriptsize{$\pm$3.34} & 69.76 \\
        
        \textbf{\M} & $\checkmark$ & & $\checkmark$ & 
        \textbf{64.16\scriptsize{$\pm$0.33}} & \textbf{72.19\scriptsize{$\pm$0.59}} & \textbf{63.05\scriptsize{$\pm$0.53}} & \textbf{66.47} & 
        \textbf{80.37\scriptsize{$\pm$0.74}} & \textbf{78.07\scriptsize{$\pm$0.75}}& \textbf{51.93\scriptsize{$\pm$1.31}} & \textbf{70.12} \\
        \midrule
                
        ERM &  & $\checkmark$ & $\checkmark$ 
        & 53.56\scriptsize{$\pm$1.53} & 59.14\scriptsize{$\pm$1.54} & 56.71\scriptsize{$\pm$0.94} & 56.47
        & 54.17\scriptsize{$\pm$2.28} & 60.56\scriptsize{$\pm$3.26} & 42.19\scriptsize{$\pm$2.70} & 52.31 \\
        RNA-Net &  & $\checkmark$ & $\checkmark$ 
        & 51.50\scriptsize{$\pm$1.26} & 60.35\scriptsize{$\pm$1.49} & 56.86\scriptsize{$\pm$1.38} & 56.24
        & 54.91\scriptsize{$\pm$2.13} & 59.82\scriptsize{$\pm$2.13} & 43.23\scriptsize{$\pm$2.13} & 52.65 \\
        SimMMDG &  & $\checkmark$ & $\checkmark$ 
        & 57.30\scriptsize{$\pm$0.89} & 65.39\scriptsize{$\pm$1.01} & 56.19\scriptsize{$\pm$1.31} & 59.63
        & 55.61\scriptsize{$\pm$1.70} & 63.43\scriptsize{$\pm$1.57} & 45.31\scriptsize{$\pm$0.20} & 54.78 \\
        MOOSA &  & $\checkmark$ & $\checkmark$ 
        & 53.72\scriptsize{$\pm$1.41} & 67.20\scriptsize{$\pm$1.54} & 60.78\scriptsize{$\pm$2.03} & 60.57
        & 56.79\scriptsize{$\pm$1.28} & 63.58\scriptsize{$\pm$0.48} & 42.59\scriptsize{$\pm$1.76} & 54.32 \\
        CMRF &  & $\checkmark$ & $\checkmark$ 
        & 59.13\scriptsize{$\pm$0.87} & 65.08\scriptsize{$\pm$0.99} & 61.30\scriptsize{$\pm$0.28} & 61.84
        & 60.49\scriptsize{$\pm$0.46} & 64.75\scriptsize{$\pm$0.32} & 47.98\scriptsize{$\pm$1.28} & 57.74 \\
        
        \textbf{\M} &  & $\checkmark$ & $\checkmark$ & 
        \textbf{59.25\scriptsize{$\pm$0.45}} & \textbf{68.05\scriptsize{$\pm$0.59}} & \textbf{61.95\scriptsize{$\pm$0.27}} & \textbf{63.08} & 
        \textbf{63.76\scriptsize{$\pm$0.91}} & \textbf{65.64\scriptsize{$\pm$0.69}} & \textbf{49.05\scriptsize{$\pm$0.19}} & \textbf{59.48} \\
        % \hline
        \midrule
        
        ERM & $\checkmark$ & $\checkmark$ & $\checkmark$ 
        & 49.46\scriptsize{$\pm$0.83} & 53.70\scriptsize{$\pm$1.78} & 48.50\scriptsize{$\pm$3.91} & 50.55
        & 74.67\scriptsize{$\pm$3.33} & 68.54\scriptsize{$\pm$0.41} & 45.40\scriptsize{$\pm$3.58} & 62.87 \\
        RNA-Net & $\checkmark$ & $\checkmark$ & $\checkmark$ 
        & 47.12\scriptsize{$\pm$0.29} & 55.80\scriptsize{$\pm$0.69} & 50.63\scriptsize{$\pm$3.39} & 51.18
        & 76.02\scriptsize{$\pm$3.19} & 73.11\scriptsize{$\pm$3.19} & 46.42\scriptsize{$\pm$3.19} & 65.18 \\
        SimMMDG & $\checkmark$ & $\checkmark$ & $\checkmark$ 
        & 60.31\scriptsize{$\pm$1.96} & 68.75\scriptsize{$\pm$1.09} & 60.97\scriptsize{$\pm$1.01} & 63.34
        & 76.38\scriptsize{$\pm$0.65} & 74.36\scriptsize{$\pm$1.70} & 51.59\scriptsize{$\pm$2.50} & 67.44 \\
        MOOSA & $\checkmark$ & $\checkmark$ & $\checkmark$ 
        & 60.00\scriptsize{$\pm$0.19} & 67.38\scriptsize{$\pm$1.07} & 64.24\scriptsize{$\pm$5.15} & 63.87
        & 75.34\scriptsize{$\pm$3.11} & 74.54\scriptsize{$\pm$1.85} & 48.44\scriptsize{$\pm$4.82} & 66.11 \\
        CMRF & $\checkmark$ & $\checkmark$ & $\checkmark$ 
        & 61.86\scriptsize{$\pm$0.74} & 69.58\scriptsize{$\pm$0.39} & 64.97\scriptsize{$\pm$0.14} & 65.47
        & 78.25\scriptsize{$\pm$0.43} & 78.07\scriptsize{$\pm$0.50} & \textbf{55.61\scriptsize{$\pm$1.23}} & 70.64 \\
        
        \textbf{\M} & $\checkmark$ & $\checkmark$ & $\checkmark$ & 
        \textbf{63.63\scriptsize{$\pm$0.39}} & \textbf{73.18\scriptsize{$\pm$0.59}} & \textbf{65.92\scriptsize{$\pm$0.44}} & \textbf{67.58} &
        \textbf{81.28\scriptsize{$\pm$1.00}} & \textbf{78.11\scriptsize{$\pm$0.32}} & 55.48\scriptsize{$\pm$1.60} & \textbf{71.62} \\
        \bottomrule
    \end{tabular}
    }
    \caption{Multi-modal \textbf{multi-source} DG accuracy with different modalities on EPIC-Kitchens and HAC datasets. The results are averaged over 3 random seeds, with standard deviation displayed as well. The best is in \textbf{bold}.}
    \label{tab:multi_domain_DG}
\end{table*}

\subsection{Experimental Setting}
\begin{table*}[t]
    \centering
    \resizebox{\textwidth}{!}{
    \begin{tabular}{lccccccccccccccc}
        \toprule
        \multirow{4}{*}{\textbf{Method}} & \multicolumn{1}{c}{} & \multicolumn{6}{c}{\textbf{EPIC-Kitchens}} & \multirow{4}{*}{Avg.} & \multicolumn{6}{c}{\textbf{HAC}} & \multirow{4}{*}{Avg.} \\
        \cmidrule(r){3-8} \cmidrule(r){10-15}
        & \multicolumn{1}{c}{S:} & \multicolumn{2}{c}{D1} & \multicolumn{2}{c}{D2} & \multicolumn{2}{c}{D3} & \multicolumn{1}{c}{} & \multicolumn{2}{c}{H} & \multicolumn{2}{c}{A} & \multicolumn{2}{c}{C} \\
        \cmidrule(r){3-4} \cmidrule(r){5-6} \cmidrule(r){7-8} \cmidrule(r){10-11} \cmidrule(r){12-13} \cmidrule(r){14-15}
        & \multicolumn{1}{c}{T:} & D2 & D3 & D1 & D3 & D1 & D2 & \multicolumn{1}{c}{} & A & C & H & C & H & A \\
        \midrule
        ERM & & 43.42\scriptsize{$\pm$3.08} & 41.99\scriptsize{$\pm$3.76} & 42.18\scriptsize{$\pm$1.74} & 42.24\scriptsize{$\pm$2.00} & 42.28\scriptsize{$\pm$1.80} & 51.98\scriptsize{$\pm$1.35} & 44.02 &
        57.36\scriptsize{$\pm$2.36} & 41.21\scriptsize{$\pm$1.95} & 69.38\scriptsize{$\pm$2.44} & 42.89\scriptsize{$\pm$4.93} & 56.77\scriptsize{$\pm$7.48} & 57.28\scriptsize{$\pm$2.97} & 54.15 \\
        RNA-Net & & 43.72\scriptsize{$\pm$3.54} & 41.51\scriptsize{$\pm$2.79} & 41.41\scriptsize{$\pm$0.82} & 44.56\scriptsize{$\pm$3.28} & 42.40\scriptsize{$\pm$3.67} & 53.46\scriptsize{$\pm$0.55} & 44.51 &
        57.91\scriptsize{$\pm$1.58} & 34.96\scriptsize{$\pm$5.02} & 71.14\scriptsize{$\pm$0.92} & 44.67\scriptsize{$\pm$1.46} & 63.25\scriptsize{$\pm$3.72} & 63.13\scriptsize{$\pm$3.72} & 55.84 \\
        SimMMDG & &
        54.09\scriptsize{$\pm$1.07} & 49.21\scriptsize{$\pm$1.20} & 53.56\scriptsize{$\pm$3.07} & 57.27\scriptsize{$\pm$2.02} & 54.42\scriptsize{$\pm$1.69} & 64.74\scriptsize{$\pm$1.77} & 55.55 &
        67.03\scriptsize{$\pm$2.23} & 42.74\scriptsize{$\pm$1.99} & 69.72\scriptsize{$\pm$2.97} & 46.26\scriptsize{$\pm$5.83} & 66.71\scriptsize{$\pm$3.29} & 65.82\scriptsize{$\pm$4.26} & 59.71 \\
        MOOSA & &
        54.99\scriptsize{$\pm$0.76} & 49.76\scriptsize{$\pm$1.57} & 52.83\scriptsize{$\pm$1.35} & 55.15\scriptsize{$\pm$2.59} & 55.49\scriptsize{$\pm$2.23} & 65.27\scriptsize{$\pm$1.06} & 55.58 &
        62.21\scriptsize{$\pm$2.40} & 43.41\scriptsize{$\pm$2.48} & 73.78\scriptsize{$\pm$0.79} & \textbf{51.23\scriptsize{$\pm$3.56}} & 63.90\scriptsize{$\pm$2.75} & 68.18\scriptsize{$\pm$2.37} & 60.45 \\
        CMRF & &
        59.35\scriptsize{$\pm$0.24} & 54.38\scriptsize{$\pm$0.32} & 57.62\scriptsize{$\pm$0.37} & 60.96\scriptsize{$\pm$0.30} & 57.01\scriptsize{$\pm$0.32} & 67.72\scriptsize{$\pm$0.61} & 59.51 &
        67.84\scriptsize{$\pm$0.96} & 46.26\scriptsize{$\pm$1.82} & 74.04\scriptsize{$\pm$0.44} & 49.36\scriptsize{$\pm$0.40} & 65.66\scriptsize{$\pm$1.58} & 65.89\scriptsize{$\pm$1.00} & 61.51 \\
        \M & &
        \textbf{59.50\scriptsize{$\pm$0.60}} & \textbf{54.89\scriptsize{$\pm$0.93}} & \textbf{58.11\scriptsize{$\pm$0.84}} & \textbf{61.73\scriptsize{$\pm$0.45}} & \textbf{57.18\scriptsize{$\pm$0.65}} & \textbf{70.77\scriptsize{$\pm$0.44}} & \textbf{60.36} &
        \textbf{69.24\scriptsize{$\pm$0.75}} & \textbf{46.81\scriptsize{$\pm$0.68}} & \textbf{76.16\scriptsize{$\pm$0.07}} & 49.36\scriptsize{$\pm$0.26} & \textbf{73.37\scriptsize{$\pm$1.12}} & \textbf{71.93\scriptsize{$\pm$3.29}} & \textbf{64.48} \\
        \bottomrule
    \end{tabular}
    }
    \caption{Multi-modal \textbf{single-source} DG accuracy with video, audio and flow modalities on EPIC-Kitchens and HAC datasets. S denotes source domain while T denotes target domain. The results are averaged over 3 random seeds, with standard deviation displayed as well. The best is in \textbf{bold}.}
    \label{tab:single_domain_DG}
\end{table*}

\textbf{Dataset \& Implementation Details.}
We conduct experiments on two benchmark datasets: EPIC-Kitchens \cite{damen2020epic} and Human-Animal-Cartoon (HAC) \cite{dong2023simmmdg}, both containing video, optical flow, and audio modalities. For EPIC-Kitchens, we adopt domains D1, D2, and D3, while for HAC, the domains are human (H), animal (A), and cartoon (C). Our experimental protocol follows the setup in \cite{dong2023simmmdg}. 
% Additional details regarding model architectures, hyperparameters, and the training environment are provided in the Appendix.

\noindent\textbf{Baselines.}
We compare our method, \M, against five baselines: ERM (a naive multi-modal approach based on feature concatenation) and four state-of-the-art MMDG methods\textemdash RNA-Net \cite{planamente2022domain}, SimMMDG \cite{dong2023simmmdg}, MOOSA \cite{dong2024towards}, and CMRF \cite{fan2024cross}. For each method, we evaluate the model achieving the best validation (in-domain) performance on the test set (out-of-domain). All results are reported as Top-1 accuracy, averaged over 3 random seeds.

\subsection{Main Results}
\textbf{Multi-Modal Multi-Source DG.}
Table~\ref{tab:multi_domain_DG} reports the performance of our method \M~and several baselines on the EPIC-Kitchens and HAC datasets under the multi-modal, multi-source domain generalization setting, where models are trained on multiple source domains and evaluated on a distinct target domain. We assess generalization by testing all pairwise modality combinations as well as the full tri-modal setup. As shown in Table~\ref{tab:multi_domain_DG}, \M~consistently surpasses baseline methods across nearly all configurations, achieving up to a 1.9\% improvement in average accuracy. In particular, utilizing all three modalities leads to the best performance, outperforming any two-modality combination. This demonstrates that \M~effectively mitigates modality imbalance and fully exploits complementary information across modalities, whereas baseline approaches struggle to benefit from the inclusion of additional modalities.

\noindent\textbf{Multi-Modal Single-Source DG.}
% We further evaluate the generalization ability of our \M~in a more challenging single-source DG setting, where models are trained using data from only one source domain and tested on multiple unseen target domains. The results using all three modalities are shown in Table~\ref{tab:single_domain_DG}. Even under this limited training scenario, our method outperforms all baselines in terms of average Top-1 accuracy, demonstrating its robustness in extracting transferable multi-modal representations from a single domain. Further experimental results involving different combinations of modalities can be found in the Appendix.
We further evaluate the generalization ability of our \M~framework in a more challenging single-source DG setting, where models are trained using data from only one source domain and tested on multiple unseen target domains. This setting imposes stronger constraints on model robustness. The results using all three modalities are summarized in Table~\ref{tab:single_domain_DG}. Even under this limited training scenario, our method consistently outperforms all baselines in terms of average Top-1 accuracy across target domains, demonstrating its strong ability to extract transferable and complementary multi-modal representations from a single domain. 

\begin{table}[t]
    \centering
    \resizebox{\linewidth}{!}{
    \begin{tabular}{lcccc}
        \toprule
        \multirow{2.5}{*}{\textbf{Method}} & \multicolumn{4}{c}{\textbf{EPIC-Kitchens}} \\
        \cmidrule(r){2-4} \multicolumn{1}{c}{} & \multicolumn{1}{c}{D2,D3} & \multicolumn{1}{c}{D1,D3} & \multicolumn{1}{c}{D1,D2} & \multicolumn{1}{c}{Avg.} \\
        \midrule
        % ERM & \\
        % RNA-Net & \\
        SimMMDG & 82.19\scriptsize{$\pm$0.39} & 78.80\scriptsize{$\pm$0.26} & 78.54\scriptsize{$\pm$1.04} & 79.84 \\
        MOOSA & 82.68\scriptsize{$\pm$0.53} & 78.64\scriptsize{$\pm$0.35} & 77.10\scriptsize{$\pm$0.42} & 79.47 \\
        CMRF & 83.22\scriptsize{$\pm$0.10} & 77.95\scriptsize{$\pm$0.24} & 78.28\scriptsize{$\pm$0.38} & 79.82 \\
        \M & \textbf{84.76\scriptsize{$\pm$0.27}} & \textbf{80.20\scriptsize{$\pm$0.35}} & \textbf{79.83\scriptsize{$\pm$0.25}} & \textbf{81.60} \\
        \toprule
    \end{tabular}}
    \caption{Multi-modal \textbf{in-domain} Accuracy on EPIC-Kitchens.}
    \label{tab:in_domain}
\end{table}
\noindent\textbf{Multi-Modal In-Domain Performance.}
In addition to strong generalization ability in MMDG settings, \M~also achieves the best performance in the in-domain scenario. As shown in Table~\ref{tab:in_domain}, when evaluated within the EPIC-Kitchens dataset under the standard in-domain setting\textemdash where training and testing domains are drawn from the same distribution\textemdash \M~consistently outperforms existing state-of-the-art baselines across all domain splits. This superior in-domain performance demonstrates the effectiveness of our model not only in handling domain shifts but also in fully exploiting multi-modal cues under standard conditions.
% Further experimental results involving different combinations of modalities can be found in the Appendix.

\subsection{Ablation Studies}
\textbf{Ablation on Each Module.}
\begin{table}[t]
    \centering
    \resizebox{\linewidth}{!}{
    \begin{tabular}{cccccccc}
        \toprule
        % gradient consistent enhancemen
        % modality dropout
        AMD & GCC & EMA & DL & \multicolumn{1}{c}{D2,D3$\rightarrow$D1} & \multicolumn{1}{c}{D1,D3$\rightarrow$D2} & \multicolumn{1}{c}{D1,D2$\rightarrow$D3} & Avg. \\ \midrule
        & & & & 
        53.56\scriptsize{$\pm$1.53} & 59.14\scriptsize{$\pm$1.54} & 56.71\scriptsize{$\pm$0.94} & 56.47 \\
        & & \checkmark & &
        57.52\scriptsize{$\pm$0.02} & 64.90\scriptsize{$\pm$0.50} & 60.41\scriptsize{$\pm$0.35} & 60.94 \\
        \checkmark & & \checkmark & &
        58.63\scriptsize{$\pm$0.16} & 65.57\scriptsize{$\pm$0.70} & 60.58\scriptsize{$\pm$0.67} & 61.59 \\
        & \checkmark & \checkmark & & 58.46\scriptsize{$\pm$1.08} & 67.27\scriptsize{$\pm$0.61} & 61.10\scriptsize{$\pm$0.32} & 62.28 \\
        & & \checkmark & \checkmark & 
        58.81\scriptsize{$\pm$0.14} & 66.52\scriptsize{$\pm$1.23} & \textbf{62.22\scriptsize{$\pm$0.41}} & 62.52 \\
        \checkmark & \checkmark & \checkmark & \checkmark &
        \textbf{59.25\scriptsize{$\pm$0.45}} & \textbf{68.05\scriptsize{$\pm$0.59}} & 61.95\scriptsize{$\pm$0.27} & \textbf{63.08} \\
        \bottomrule
    \end{tabular}
    }
    \caption{Abalation on each design on EPIC-Kitchens with flow and audio data. AMD: adaptive modality dropout, GCC: gradient consistency constraint, EMA: exponential moving average, DL: distillation loss.}
    \label{tab:ablation_on_each_design}
\end{table}
We conducted ablation studies on EPIC-Kitchens using flow and audio modalities to assess the contribution of each component in our framework. As shown in Table~\ref{tab:ablation_on_each_design}, introducing EMA alone leads to a substantial improvement over the baseline (from 56.47\% to 60.94\%), highlighting the effectiveness of temporal knowledge transfer. Adding adaptive modality dropout (AMD) or the gradient consistency constraint (GCC) individually further boosts performance, suggesting that both dynamic modality regulation and uni-modal enhancement play important roles in promoting robust multi-modal learning. Incorporating distillation loss (DL) on top of EMA also yields notable gains. Finally, combining all modules achieves the best average performance , demonstrating their complementary benefits and the overall effectiveness of our proposed design.

% \noindent\textbf{Ablation on mutual improvement}

\noindent\textbf{Ablation on Modality Drop.}
\begin{table}[t]
    \centering
    \resizebox{\linewidth}{!}{
    \begin{tabular}{ccccc}
        \toprule
        Method & \multicolumn{1}{c}{A,C$\rightarrow$H} & \multicolumn{1}{c}{H,C$\rightarrow$A} & \multicolumn{1}{c}{H,A$\rightarrow$C} & Avg. \\ 
        w/o dropout & 62.82\scriptsize{$\pm$1.77} & 64.24\scriptsize{$\pm$1.26} & 45.22\scriptsize{$\pm$0.67} & 57.43 \\
        fix dropout & \textbf{64.34\scriptsize{$\pm$0.65}} & 64.42\scriptsize{$\pm$0.59} & 47.89\scriptsize{$\pm$0.69} & 58.88 \\
        adaptive dropout & 63.76\scriptsize{$\pm$0.91} & \textbf{65.64\scriptsize{$\pm$0.69}} & \textbf{49.05\scriptsize{$\pm$0.19}} & \textbf{59.48} \\
        \bottomrule
    \end{tabular}
    }
    \caption{Comparison with different dropout strategy on HAC with flow and audio data.}
    \label{tab:ablation_on_mdality_drop}
\end{table}
We compare our adaptive modality dropout with two variants: (1) w/o dropout, where no modality is dropped during fusion; (2) fix dropout, where modalities are dropped with a fixed probability of 0.5.
As shown in Table~\ref{tab:ablation_on_mdality_drop}, all dropout strategies improve over the baseline without dropout, indicating the benefit of suppressing modality dominance. Among them, our adaptive dropout achieves the best overall performance, demonstrating its effectiveness in dynamically balancing modality contributions based on training feedback.

\subsection{Further Analysis}
\begin{figure}[t]
    \centering
    \includegraphics[width=\linewidth]{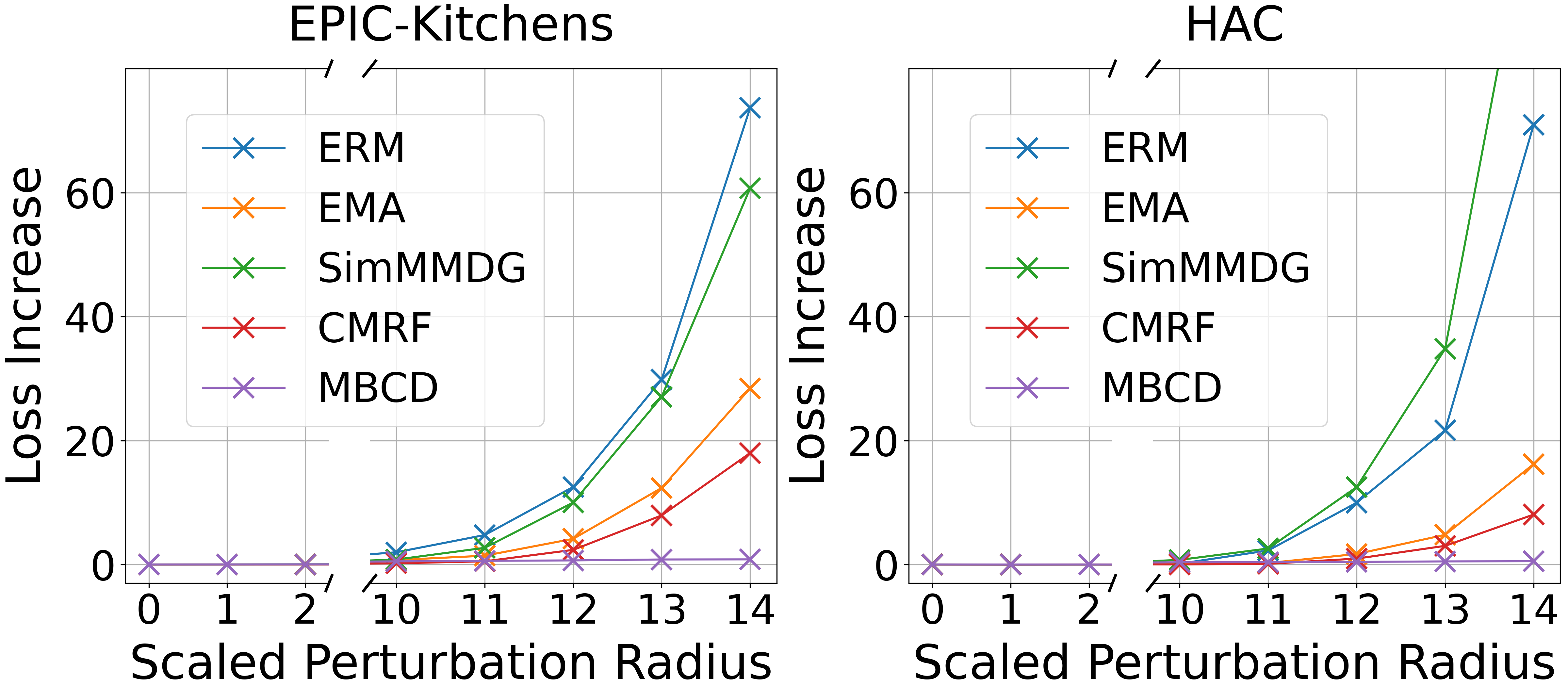}
    \caption{Comparison of flatness for different methods on EPIC-Kitchens and HAC.}
    \label{fig:flatness}
\end{figure}

\noindent\textbf{Flatness Analysis.}
To evaluate the flatness of the loss landscape, we follow prior work \cite{cha2021swad} and measure the change in loss values under perturbations around the converged parameters. The core assumption is that flatter minima are indicative of better generalization. Specifically, we add random directional noise by a perturbation radius and measure the corresponding loss increase. As shown in Figure~\ref{fig:flatness}, the x-axis represents the scaled perturbation radius (magnified 10$\times$), and the y-axis shows the corresponding loss increase.
Our results reveal that \M~consistently converges to flatter minima than all baselines across both EPIC-Kitchens (D1, D3 $\rightarrow$ D2) and HAC (A, C $\rightarrow$ H) datasets. Furthermore, other WA methods like EMA and CMRF also yield flatter solutions compared to non-averaging methods, underscoring the general benefit of WA for discovering flat minima.

\noindent\textbf{Cross-Modal Interaction Analysis.}
\begin{figure}[t]
    \centering
    \includegraphics[width=\linewidth]{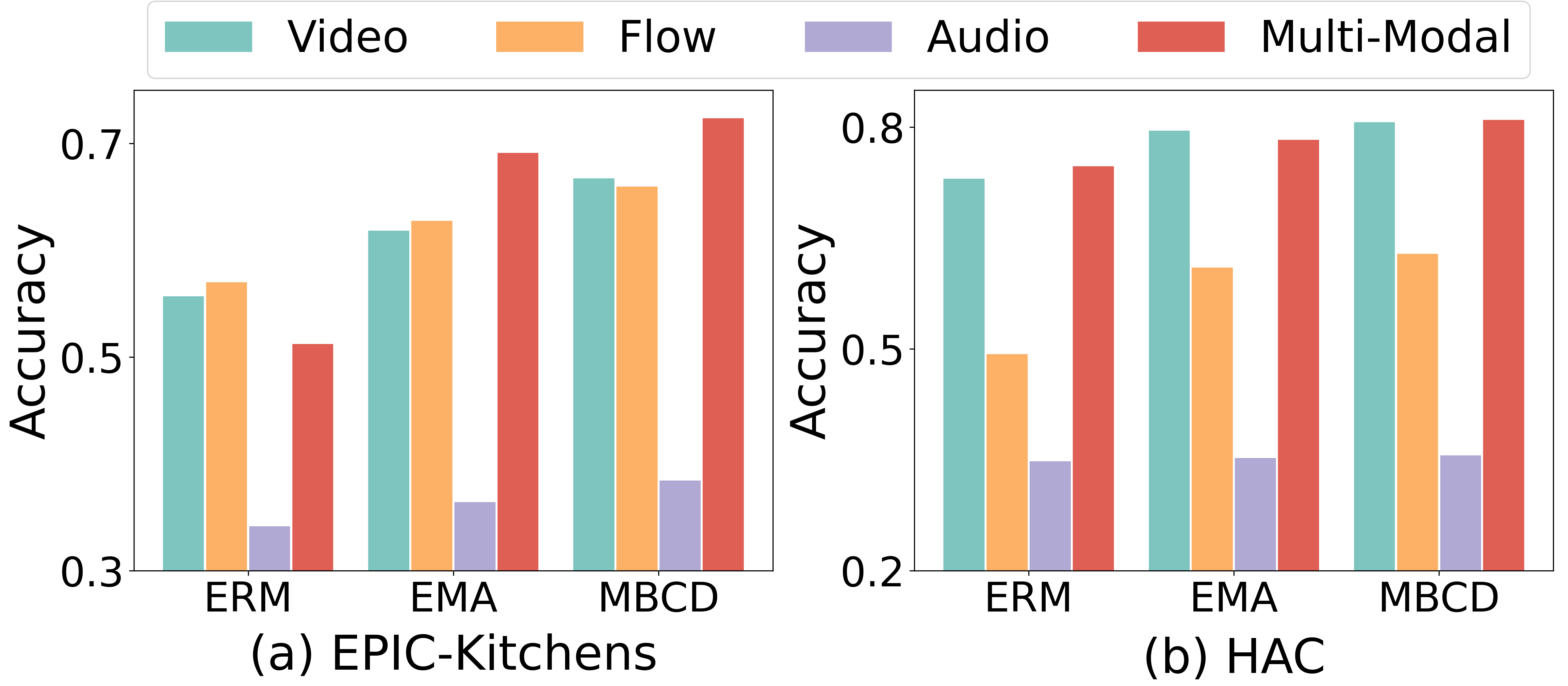}
    \caption{Comparison of modality-wise and fused accuracies on
    EPIC-Kitchens and HAC.}
    \label{fig:cross_modal_interaction}
\end{figure}
To demonstrate the effectiveness of \M~in enhancing cross-modal interaction, we compare modality-wise and fused accuracies across different methods. As shown in Figure~\ref{fig:cross_modal_interaction}, \M~consistently improves the performance of all individual modalities and achieves higher overall accuracy on both EPIC-Kitchens (D1, D3 $\rightarrow$ D2) and HAC (A, C $\rightarrow$ H). Compared to \M, EMA achieves slightly lower performance, while ERM performs the worst\textemdash primarily due to its imbalanced optimization. Under standard ERM training, multi-modal fusion often suffers from imbalanced learning, leading to the underutilization of weaker modalities like audio and causing the fused model to underperform. In contrast, \M~promotes balanced optimization across modalities. This not only enhances the representational quality of each modality but also strengthens their cross-modal complementarity, ultimately leading to consistently superior fusion performance.

\noindent\textbf{Training Stability Analysis.}
\begin{figure}[t]
    \centering
    \includegraphics[width=\linewidth]{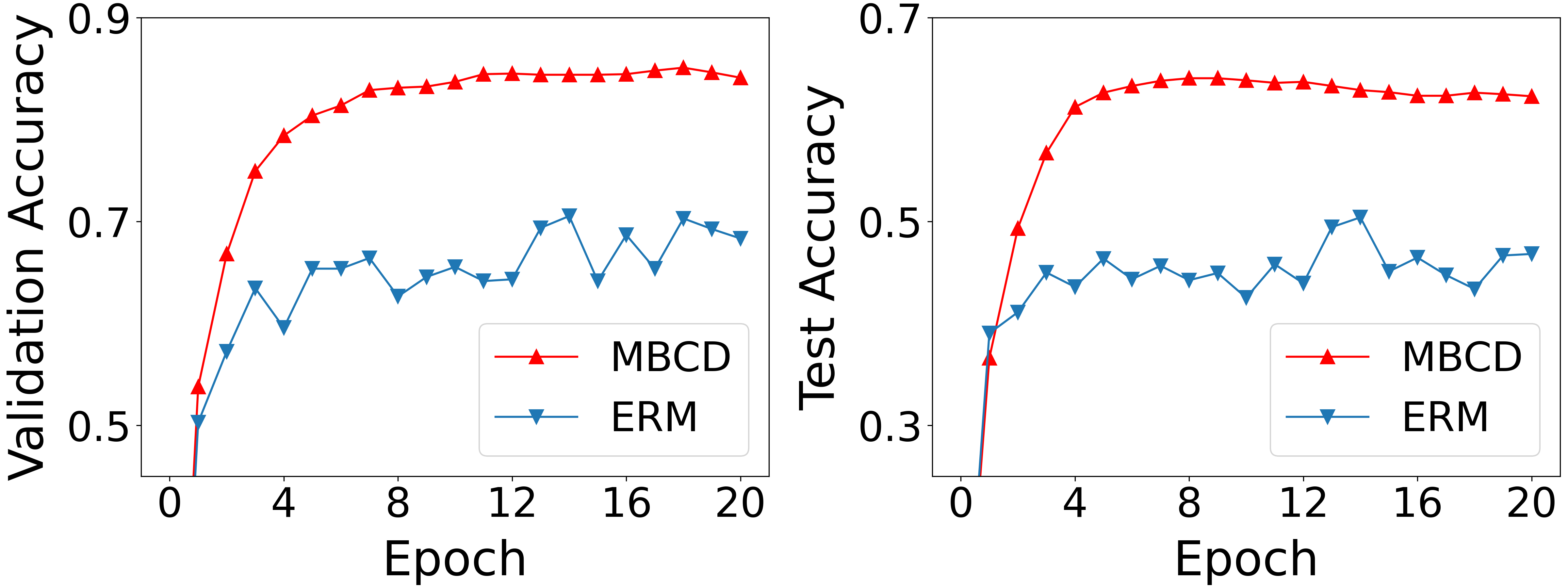}
    \caption{Validation and test accuracy curves on EPIC-Kitchens (D2,D3$\rightarrow$D1) across all modalities.}
    \label{fig:stability}
\end{figure}
% We examine the robustness of out-of-domain performance for \M~when model selection is guided by an in-domain validation set. As shown in Figure~\ref{fig:stability}, \M~exhibits remarkable training stability, a critical attribute for reliable deployment. This stability is evident from two perspectives. The first is the smoothness of both the validation and test accuracy curves across all training epochs, which points to a highly stable optimization process with low variance. The second is the tight correlation between validation and test accuracy curves, which demonstrates consistent generalization and a robust defense against overfitting. This strong alignment is a testament to \M's ability to generalize reliably under domain shifts. In contrast, the ERM baseline exhibits significantly unstable performance, highlighting \M's superiority in both convergence and generalization.
We examine the robustness of out-of-domain performance for \M~when model selection is guided by an in-domain validation set. As shown in Figure~d\ref{fig:stability}, \M~exhibits remarkable training stability, a critical attribute for reliable deployment in real-world scenarios. This stability is evident from two perspectives. The first is the smoothness of both the validation and test accuracy curves across all training epochs, which points to a highly stable optimization process with low variance and minimal fluctuations. The second is the tight correlation between validation and test accuracy curves, which demonstrates consistent generalization and a robust defense against overfitting, even under significant distributional shifts. This strong alignment is a testament to \M's ability to generalize reliably under domain shifts. In contrast, the ERM baseline exhibits significantly unstable performance, highlighting \M's clear superiority in both convergence dynamics and generalization ability.

\noindent\textbf{Visualization of Embeddings.}
\begin{figure}[t]
    \centering
    \includegraphics[width=\linewidth]{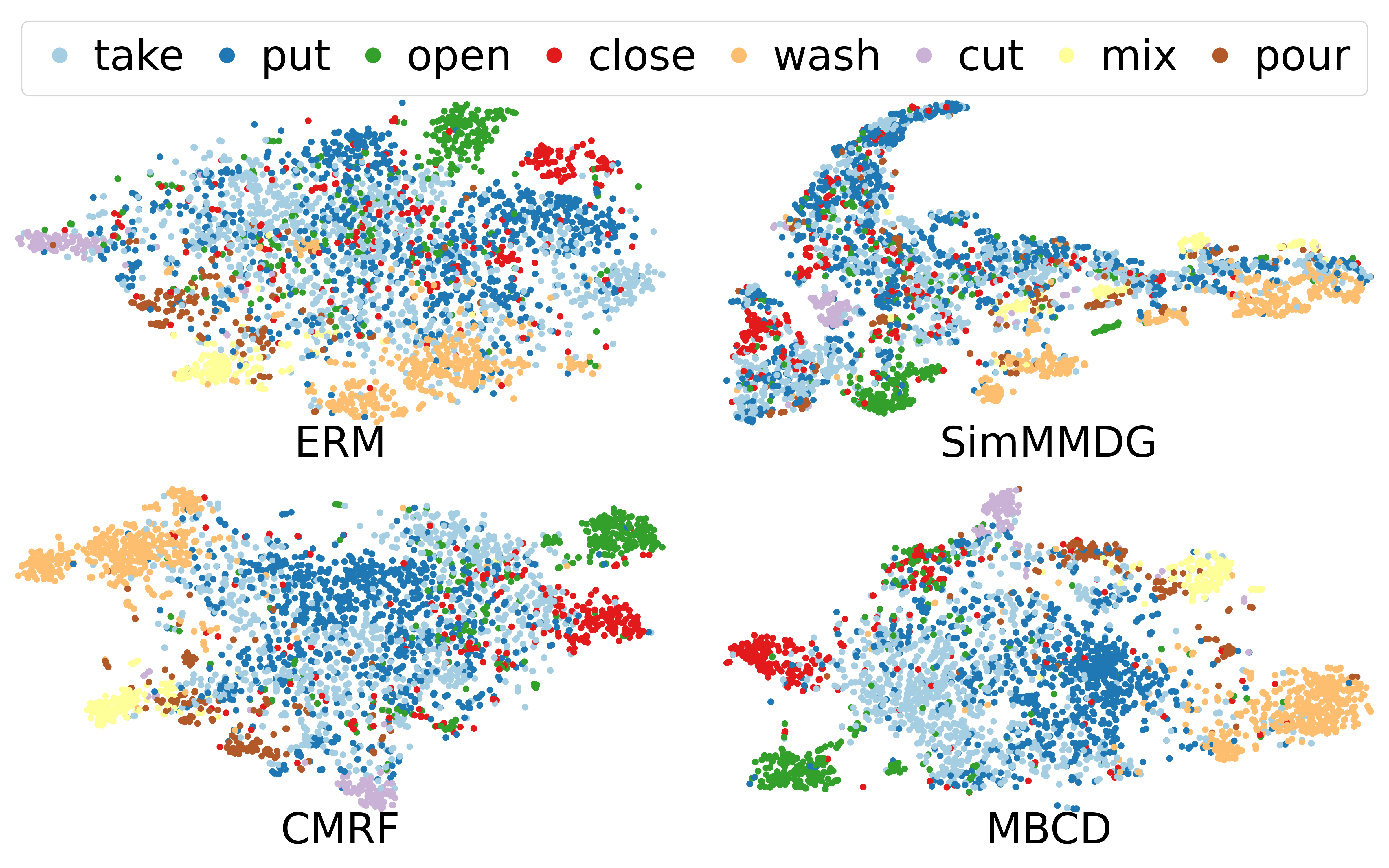}
    \caption{T-SNE visualization of concatenated multi-modal embeddings on EPIC-Kitchens.}
    \label{fig:distribution}
\end{figure}
To investigate the quality of the learned representations and gain insights into our model's mechanism under domain shift, we visualize the concatenated multi-modal embeddings of the out-of-domain testing set using t-SNE \cite{maaten2008visualizing}. As depicted in Figure~\ref{fig:distribution}, we compare the distributions of \M~with those of ERM, SimMMDG, and CMRF. The results demonstrate that while the baseline methods produce highly intertwined clusters with significant class overlap, our model, \M, successfully learns distinct and well-separated action representations. Specifically, we observe that the ERM and SimMMDG baselines struggle to differentiate between action types, leading to a blurry and mixed distribution of embeddings. While CMRF shows some improvement, it still exhibits a noticeable confusion between semantically similar actions, such as 'take' and 'put'. In contrast, \M~successfully disentangles these action patterns, forming compact and isolated clusters for each class. This visual evidence confirms that \M~learns a more structured and semantically meaningful embedding space, which is crucial for its robust generalization to unseen domains.

%% file: 5-conclusion.tex
In this work, we analyzed the core limitation of applying EMA to MMDG and proposed \M, a unified framework that preserved the flatness-inducing properties of WA while overcoming its limitations in multi-modal settings. By leveraging collaborative distillation, \M~promoted balanced optimization across modalities and guided the model toward flatter, more generalizable solutions. Experiments on EPIC-Kitchens and HAC benchmarks showed that \M~consistently outperformed state-of-the-art methods across diverse modality settings. In future work, we planned to explore the theoretical foundations of WA strategies in multi-modal learning, aiming to better understand their role in MMDG and to provide deeper insights and stronger foundations for advancing this direction.

%% file: 6-acknowledgment.tex
This work was supported by National Natural Science Foundation of China (Grant No. 62572097, No. 62176043, and No. U22A2097).

%% file: 7-appendix.tex
\section{Appendix}
\subsection{Theoretical Insights}
\textbf{Gradient Consistency Analysis.}
We provide a theoretical analysis to better understand the proposed uni-modal objective-guided strategy and illustrate how it improves gradient consistency. For simplicity, we omit $\phi$ in the second term of Eq.~\ref{eq:gradient_consistency}, since the prediction head is not directly affected by the strategy under discussion. We perform a first-order Taylor expansion around $\theta_k$ as follows:
\begin{align}
    &\min_{\theta}\mathcal{L}_{cls}(\bigcup_{k=1}^{M}\theta_k';x) \nonumber \\
    = &\min_{\theta}\mathcal{L}_{cls}(\bigcup_{k=1}^{M}\theta_k-\alpha\nabla_{\theta_k}\mathcal{L}_{cls}(\theta_k;x_k);x) \label{eq:mm_cls} \\
    % = &\min_{\theta}\mathcal{L}_{cls}(\theta;x) - \sum_{k=1}^{M}\alpha\nabla_{\theta_k}\mathcal{L}_{cls}(\theta_k;x_k) \cdot \nabla_{\theta_k}\mathcal{L}_{cls}(\theta;x) \nonumber \\
    % &+ O(\alpha\nabla_{\theta_k}\mathcal{L}_{cls}(\theta_k;x_k)) \label{eq:mm_cls_expansion} \\
    \approx &\min_{\theta}\mathcal{L}_{cls}(\theta;x) - \sum_{k=1}^{M}\alpha\nabla_{\theta_k}\mathcal{L}_{cls}(\theta_k;x_k) \cdot \nabla_{\theta_k}\mathcal{L}_{cls}(\theta;x). \label{eq:mm_cls_expansion}
\end{align}

To prove the derivation from Eq.~\ref{eq:mm_cls} to Eq.~\ref{eq:mm_cls_expansion}, we first consider a multivariable function $\mathcal{L}(\theta^{\prime})$, where $\theta^{\prime} = \bigcup_{k=1}^{M} \theta_k^{\prime}$. Its first-order Taylor expansion around the point $\theta$ is given by:
\begin{align}
    \mathcal{L}(\theta^{\prime})
    \approx\mathcal{L}(\theta)
    +\nabla\mathcal{L}(\theta)^T(\theta^{\prime}-\theta)
    + \cdots, \nonumber
\end{align}
where $\nabla\mathcal{L}(\theta)$ denotes the full gradient vector, which can be decomposed into gradients with respect to each sub-parameter:
\begin{align}
    \nabla\mathcal{L}(\theta) = [\nabla_{\theta_1}\mathcal{L}(\theta), \nabla_{\theta_2}\mathcal{L}(\theta), \cdots, \nabla_{\theta_M}\mathcal{L}(\theta)]. \nonumber
\end{align}
Accordingly, the inner product term in the expansion can be expressed as:
\begin{align}
    \nabla\mathcal{L}(\theta)^T(\theta^{\prime}-\theta)
    =\sum_{k=1}^M\nabla_{\theta_k}\mathcal{L}(\theta)^T(\theta_k^{\prime}-\theta_k). \nonumber
\end{align}
Returning to our derivation, we consider the case where each $\theta_k$ is updated by a gradient descent step with respect to its own modality-specific loss:
\begin{align}
    &\min_{\theta}\mathcal{L}_{cls}(\bigcup_{k=1}^{M}\theta_k-\alpha\nabla_{\theta_k}\mathcal{L}_{cls}(\theta_k;x_k);x) \nonumber \\
    = &\min_{\theta}\mathcal{L}_{cls}(\theta;x)-\alpha[\nabla_{\theta_1}\mathcal{L}_{cls}(\theta;x), \cdots, \nabla_{\theta_M}\mathcal{L}_{cls}(\theta;x)]^T \nonumber \\
    &\cdot[\nabla_{\theta_1}\mathcal{L}_{cls}(\theta_1;x), \cdots, \nabla_{\theta_M}\mathcal{L}_{cls}(\theta_M;x)] \nonumber \\
    =&\min_{\theta}\mathcal{L}_{cls}(\theta;x) - \sum_{k=1}^{M}\alpha\nabla_{\theta_k}\mathcal{L}_{cls}(\theta_k;x_k) \cdot \nabla_{\theta_k}\mathcal{L}_{cls}(\theta;x) \nonumber.
\end{align}

Building on this derivation, the objective in Eq.~\ref{eq:gradient_consistency} can be reformulated as:
\begin{align}
    \min_{\theta,\phi}&\sum_{k=1}^M\mathcal{L}_{cls}(\theta_k,\phi_k;x_k) + \mathcal{L}_{cls}(\theta,\phi_{mm};x) - \nonumber \\ 
    &\sum_{k=1}^{M}\alpha\nabla_{\theta_k}\mathcal{L}_{cls}(\theta_k,\phi_k;x_k) \cdot \nabla_{\theta_k}\mathcal{L}_{cls}(\theta,\phi;x).
\end{align}

This Eq. reveals that the learning objective aims to minimize both the uni-modal losses $\sum_{k=1}^M\mathcal{L}_{cls}(\theta_k,\phi_k;x_k)$ and the multi-modal loss $\mathcal{L}_{cls}(\theta,\phi_{mm};x)$, while simultaneously maximizing the inner product between the gradients of uni-modal and multi-modal losses. Maximizing this inner product encourages the gradients $\nabla_{\theta_k}\mathcal{L}_{cls}(\theta_k,\phi_k;x_k)$ and $\nabla_{\theta_k}\mathcal{L}_{cls}(\theta,\phi;x)$ to be aligned, effectively promoting consistency in their update directions. In essence, this can be viewed as a form of gradient matching, which enhances coordination between uni-modal and multi-modal learning objectives.

\noindent\textbf{MMDG Analysis.}
To theoretically analyze the generalization behavior of our multi-modal framework, we adopt the classical bias–variance decomposition under the expected cross-entropy loss. For test samples drawn from unseen domain distribution $P_{\mathcal{T}}$, the expected generalization error can be decomposed as:
\begin{align}
    &\mathbb{E}_{(x, y)\sim P_{\mathcal{T}}}\mathbb{E}_{\mathcal{S}}[C E(f(\mathbf{x} ;\mathcal{S}),y)] \nonumber \\
    =&\underbrace{\mathbb{E}_{(x, y)\sim P_{\mathcal{T}}}[C E(y, \bar{f}(\mathbf{x}))]}_{\text {Bias }^{2}} 
    +\underbrace{\mathbb{E}_{x}\mathbb{E}_{\mathcal{S}}[\mathrm{KL}(\bar{f}(x), f(x;\mathcal{S}))]}_{\text{Variance}}.
\end{align}

In multi-modal learning, modality imbalance amplifies the bias term. This occurs when the model becomes overly reliant on dominant modalities, causing misalignment when these modalities are degraded in target domains. Our Adaptive Modality Dropout addresses this by encouraging balanced contributions across all modalities, thereby promoting the use of complementary information and mitigating the bias.

Meanwhile, the variance term reflects the model's sensitivity to data sampling and initialization. To suppress this, we employ an EMA-based teacher, which provides a smoothed target for cross-modal distillation. This temporal ensembling technique stabilizes the training dynamics and reduces prediction variability, ultimately enhancing the model's robustness to distribution shifts.

\subsection{Experimental Setting}
\noindent\textbf{Dataset.}
We evaluate our method on two widely adopted benchmark datasets: EPIC-Kitchens \cite{damen2020epic} and Human-Animal-Cartoon (HAC) \cite{dong2023simmmdg}, following the experimental protocols outlined in their respective original works.
EPIC-Kitchens comprises eight action categories—‘put’, ‘take’, ‘open’, ‘close’, ‘wash’, ‘cut’, ‘mix’, and ‘pour’—captured in three distinct kitchen environments, denoted as domains D1, D2, and D3.
HAC features seven action classes—‘sleeping’, ‘watching TV’, ‘eating’, ‘drinking’, ‘swimming’, ‘running’, and ‘opening door’—performed by three entity types: humans (H), animals (A), and cartoon characters (C), which correspond to domains H, A, and C. 

Both datasets provide three temporally aligned modalities: RGB video, audio, and optical flow. The number of action segments per domain is detailed in Table~\ref{tab:domain_count}. For each domain, the data are partitioned into a training set and a validation set. When a domain is used as a source, its training subset is employed for model optimization, while the validation subset is used for early stopping. In contrast, when a domain is designated as the target, the full set—including both training and validation data—is utilized solely for evaluation purposes. \\
\begin{table}[ht]
    \centering
    \resizebox{\linewidth}{!}{
    \begin{tabular}{lccccccc}
        \toprule
        \multirow{2.5}{*}{\textbf{Mode}} & & \multicolumn{3}{c}{\textbf{EPIC-Kitchens}} & \multicolumn{3}{c}{\textbf{HAC}} \\
        \cmidrule(r){3-5} \cmidrule(r){6-8}
        \multicolumn{1}{c}{} & \multicolumn{1}{c}{\textbf{Domain}} & D1 & D2 & D3 & H & A & C \\
        \midrule
        \textbf{Training} & & 1543 & 2495 & 3897 & 1111 & 730 & 870 \\ 
        \textbf{Validate} & & 435 & 750 & 974 & 276 & 176 & 218 \\ 
        \textbf{Testing} & & 1978 & 3245 & 4871 & 1387 & 906 & 1088 \\ 
        \toprule
    \end{tabular}}
    \caption{Number of action segments per domain.}
    \label{tab:domain_count}
\end{table}

\noindent\textbf{Baseline.} We provide a detailed description of the baselines used in our evaluation.
\begin{itemize}
    \item \textbf{ERM} serves as a fundamental baseline, using a simple multi-modal concatenation-based training without any DG techniques.
    \item \textbf{RNA-Net} proposes a cross-modal loss to rebalance the mean feature norms of different modalities, directly addressing modality imbalance during training.
    \item \textbf{SimMMDG} decomposes features into modality-specific and modality-shared components. It employs supervised contrastive learning on shared features and distance constraints on specific features, further regularized by a cross-modal translation module to enhance generalization.
    \item \textbf{MOOSA} improves generalization through the design of two innovative multi-modal self-supervised pretext tasks, namely Masked Cross-modal Translation and Multi-modal Jigsaw Puzzles.
    \item \textbf{CMRF} builds a consistent and flat loss landscape in the shared representation space. It achieves this by creating cross-modal interpolations from a moving averaged teacher model and using feature distillation to optimize high-loss regions between modalities.
\end{itemize}

\noindent\textbf{Additional Implementation Details.} 
To ensure fair comparison with prior work, we adopt the same modality-specific backbones as used in SimMMDG \cite{dong2023simmmdg} and \cite{fan2024cross}, and follow the implementation protocols described in \cite{damen2020epic}. Experiments are conducted across three modalities: RGB video, audio, and optical flow. Our implementation is based on the MMAction2 framework \cite{2020mmaction2}. For visual representation, we employ the SlowFast architecture \cite{feichtenhofer2019slowfast}, initialized with Kinetics-400 pre-trained weights. The audio modality is encoded using a ResNet-18 \cite{he2016deep} backbone, initialized from the VGGSound checkpoint \cite{chen2020vggsound}. For optical flow, we utilize the slow pathway of the SlowFast network, also initialized with Kinetics-400 pre-trained weights. 

The dimensionality of the uni-modal features is set to 2304 for the video modality, 512 for audio, and 2048 for optical flow, respectively. We optimize all models using the Adam optimizer \cite{kingma2014adam} with a learning rate of 0.0001 and a batch size of 16. The distillation loss coefficient $\lambda$ is fixed at 1.0 throughout all experiments. For the gradient consistency objective defined in Eq.\ref{eq:gradient_consistency}, we set the inner-loop learning rate $\alpha$ to 0.0001. The EMA smoothing coefficient $\beta$ in Eq.\ref{eq:ema} is set to 0.999, following common practice to stabilize parameter updates.

All experiments are conducted on a machine equipped with an NVIDIA GeForce RTX 4090 D GPU and an Intel(R) Core(TM) i9-14900KF CPU. Each model is trained for 20 epochs, with a total training time of approximately three hours. To ensure statistical robustness, all reported results are averaged over three runs using different random seeds, and we report both the mean and standard deviation.

\subsection{More Results}
\noindent\textbf{Sensitivity Analysis of EMA $\beta$.}
We investigate the influence of the smoothing factor $\beta$ in the EMA update, which controls the temporal decay rate of historical model weights. As shown in Tab.~\ref{tab:ablation_on_beta}, we evaluate the model performance on EPIC-Kitchens under different values of $\beta$. 

We observe that increasing $\beta$ from 0.9 to 0.999 leads to a steady improvement in average performance (from 64.66 to 67.58), suggesting that stronger smoothing reduces variance and enhances generalization. However, an excessively large $\beta$ (e.g., 0.9999) causes a sharp performance drop, likely because the EMA teacher updates too slowly at early stages, retaining poorly trained weights and providing suboptimal guidance. These results underscore the need to balance stability and adaptability when choosing $\beta$: small values cause noisy updates, while overly large ones hinder convergence.
\begin{table}[t]
    \centering
    \resizebox{\linewidth}{!}{
    \begin{tabular}{ccccc}
        \toprule
        $\beta$ & {D2,D3$\rightarrow$D1} & D1,D3$\rightarrow$D2 & D1,D2$\rightarrow$D3 & Avg. \\ 
        0.9 & 61.88\scriptsize{$\pm$1.51} & 69.93\scriptsize{$\pm$0.99} & 62.18\scriptsize{$\pm$0.33} & 64.66 \\
        0.99 & 62.76\scriptsize{$\pm$0.09} & 71.54\scriptsize{$\pm$0.61} & 64.35\scriptsize{$\pm$0.81} & 66.22 \\
        0.999 & 
        \textbf{63.63\scriptsize{$\pm$0.39}} & \textbf{73.18\scriptsize{$\pm$0.59}} & \textbf{65.92\scriptsize{$\pm$0.44}} & \textbf{67.58} \\
        0.9999 & 57.48\scriptsize{$\pm$0.34} & 63.71\scriptsize{$\pm$1.72} & 54.98\scriptsize{$\pm$2.40} & 58.72 \\
        \bottomrule
    \end{tabular}
    }
    \caption{Impact of the EMA smoothing coefficient $\beta$ on MMDG performance evaluated on EPIC-Kitchens.}
    \label{tab:ablation_on_beta}
\end{table}

\begin{table*}[!t]
    \centering
    \resizebox{\textwidth}{!}{
    \begin{tabular}{lccccccccccc}
        \toprule
        \multirow{4}{*}{\textbf{Method}} & \multicolumn{4}{c}{} & \multicolumn{6}{c}{\textbf{EPIC-Kitchens}} & \multirow{4}{*}{Avg.} \\
        \cmidrule(r){6-11}
        & \multicolumn{3}{c}{\textbf{Modality}} & \multicolumn{1}{c}{Source:} & \multicolumn{2}{c}{D1} &\multicolumn{2}{c}{D2} & \multicolumn{2}{c}{D3} \\
        \cmidrule(r){2-4} \cmidrule(r){6-7} \cmidrule(r){8-9} \cmidrule(r){10-11}
        & \multicolumn{1}{c}{Video} & \multicolumn{1}{c}{Audio} & \multicolumn{1}{c}{Flow} & \multicolumn{1}{c}{Target:} & D2 & D3 & D1 & D3 & D1 & D2 & \\
        \midrule
        ERM & $\checkmark$ & $\checkmark$ & & &
        50.24\scriptsize{$\pm$1.79} & 50.50\scriptsize{$\pm$1.10} & 45.25\scriptsize{$\pm$0.57} & 52.67\scriptsize{$\pm$3.12} & 48.11\scriptsize{$\pm$0.25} & 57.67\scriptsize{$\pm$2.01} & 50.74 \\
        RNA-Net & $\checkmark$ & $\checkmark$ & & &
        50.05\scriptsize{$\pm$2.64} & 50.77\scriptsize{$\pm$1.04} & 45.92\scriptsize{$\pm$0.91} & 53.95\scriptsize{$\pm$0.92} & 48.08\scriptsize{$\pm$0.34} & 56.56\scriptsize{$\pm$3.04} & 50.89 \\
        SimMMDG & $\checkmark$ & $\checkmark$ & & &
        49.23\scriptsize{$\pm$1.56} & 44.54\scriptsize{$\pm$3.77} & 48.55\scriptsize{$\pm$0.66} & 55.62\scriptsize{$\pm$0.97} & 48.28\scriptsize{$\pm$1.50} & 59.50\scriptsize{$\pm$2.00} & 50.95 \\
        MOOSA & $\checkmark$ & $\checkmark$ & & &
        48.58\scriptsize{$\pm$1.50} & 48.66\scriptsize{$\pm$1.23} & 49.22\scriptsize{$\pm$1.42} & 54.45\scriptsize{$\pm$1.94} & 49.48\scriptsize{$\pm$1.18} & 59.32\scriptsize{$\pm$1.45} & 51.62 \\
        CMRF & $\checkmark$ & $\checkmark$ & & &
        \textbf{54.62\scriptsize{$\pm$0.64}} & \textbf{51.63\scriptsize{$\pm$0.69}} & 53.57\scriptsize{$\pm$0.52} & 59.15\scriptsize{$\pm$0.61} & 50.57\scriptsize{$\pm$0.37} & 62.06\scriptsize{$\pm$0.35} & 55.27 \\
        \M & $\checkmark$ & $\checkmark$ & & &
        52.46\scriptsize{$\pm$2.33} & 50.21\scriptsize{$\pm$0.98} & \textbf{54.82\scriptsize{$\pm$1.20}} & \textbf{59.66\scriptsize{$\pm$1.10}} & \textbf{51.85\scriptsize{$\pm$1.08}} & \textbf{65.21\scriptsize{$\pm$1.12}} & \textbf{55.70} \\
        \midrule

        ERM & $\checkmark$ &  & $\checkmark$ & & 
        55.10\scriptsize{$\pm$0.85} & 45.97\scriptsize{$\pm$3.31} & 51.40\scriptsize{$\pm$1.67} & 54.69\scriptsize{$\pm$1.02} & 54.28\scriptsize{$\pm$2.20} & 61.41\scriptsize{$\pm$0.77} & 53.81 \\
        RNA-Net & $\checkmark$ &  & $\checkmark$ & & 
        54.20\scriptsize{$\pm$1.04} & 46.27\scriptsize{$\pm$3.09} & 51.26\scriptsize{$\pm$2.42} & 54.55\scriptsize{$\pm$1.05} & 53.40\scriptsize{$\pm$0.82} & 63.43\scriptsize{$\pm$1.35} & 53.85 \\
        SimMMDG & $\checkmark$ &  & $\checkmark$ & & 
        51.84\scriptsize{$\pm$0.94} & 44.69\scriptsize{$\pm$2.75} & 52.21\scriptsize{$\pm$0.54} & 55.85\scriptsize{$\pm$1.50} & 52.86\scriptsize{$\pm$1.12} & 62.39\scriptsize{$\pm$1.28} & 53.31 \\
        MOOSA & $\checkmark$ &  & $\checkmark$ & & 
        52.32\scriptsize{$\pm$0.60} & 47.65\scriptsize{$\pm$0.18} & 52.06\scriptsize{$\pm$1.36} & 54.60\scriptsize{$\pm$1.95} & 56.42\scriptsize{$\pm$0.39} & 65.17\scriptsize{$\pm$0.49} & 54.70 \\
        CMRF & $\checkmark$ &  & $\checkmark$ & & 
        57.02\scriptsize{$\pm$0.44} & 51.13\scriptsize{$\pm$0.84} & 56.08\scriptsize{$\pm$0.87} & 57.37\scriptsize{$\pm$0.01} & 58.61\scriptsize{$\pm$1.52} & 66.69\scriptsize{$\pm$0.25} & 57.82 \\
        \M & $\checkmark$ &  & $\checkmark$ & & 
        \textbf{58.23\scriptsize{$\pm$0.24}} & \textbf{51.66\scriptsize{$\pm$1.47}} & \textbf{59.05\scriptsize{$\pm$0.11}} & \textbf{59.01\scriptsize{$\pm$1.22}} & \textbf{59.34\scriptsize{$\pm$0.62}} & \textbf{69.18\scriptsize{$\pm$0.90}} & \textbf{59.41} \\
        \midrule

        ERM &  & $\checkmark$ & $\checkmark$  & &   
        47.93\scriptsize{$\pm$0.34} & 46.53\scriptsize{$\pm$0.45} & 45.64\scriptsize{$\pm$0.72} & 51.87\scriptsize{$\pm$0.43} & 48.33\scriptsize{$\pm$2.23} & 55.88\scriptsize{$\pm$0.94} & 49.36 \\
        RNA-Net &  & $\checkmark$ & $\checkmark$  & & 
        48.85\scriptsize{$\pm$1.15} & 48.15\scriptsize{$\pm$2.12} & 45.79\scriptsize{$\pm$0.76} & 51.74\scriptsize{$\pm$0.61} & 48.82\scriptsize{$\pm$1.01} & 57.10\scriptsize{$\pm$1.25} & 50.07 \\
        SimMMDG &  & $\checkmark$ & $\checkmark$  & & 
        51.63\scriptsize{$\pm$0.52} & 50.13\scriptsize{$\pm$1.02} & 48.52\scriptsize{$\pm$0.93} & 52.40\scriptsize{$\pm$0.84} & 54.03\scriptsize{$\pm$1.17} & 62.44\scriptsize{$\pm$1.09} & 53.19 \\
        MOOSA &  & $\checkmark$ & $\checkmark$  & & 
        47.04\scriptsize{$\pm$1.50} & 47.21\scriptsize{$\pm$0.54} & 47.99\scriptsize{$\pm$1.51} & 53.39\scriptsize{$\pm$1.08} & 54.11\scriptsize{$\pm$1.21} & 61.51\scriptsize{$\pm$1.33} & 51.88 \\
        CMRF &  & $\checkmark$ & $\checkmark$  & & 
        \textbf{54.06\scriptsize{$\pm$0.14}} & \textbf{52.21\scriptsize{$\pm$0.54}} & 52.06\scriptsize{$\pm$0.27} & 57.46\scriptsize{$\pm$0.60} & 52.92\scriptsize{$\pm$0.86} & 61.68\scriptsize{$\pm$0.78} & 55.07 \\
        \M &  & $\checkmark$ & $\checkmark$  & & 
        52.16\scriptsize{$\pm$1.11} & 49.84\scriptsize{$\pm$0.53} & \textbf{52.16\scriptsize{$\pm$0.27}} & \textbf{57.50\scriptsize{$\pm$0.46}} & \textbf{55.49\scriptsize{$\pm$0.42}} & \textbf{65.15\scriptsize{$\pm$0.61}} & \textbf{55.38} \\
        \bottomrule
    \end{tabular}}
    \caption{Multi-modal \textbf{single-source} DG accuracy with different modalities on EPIC-Kitchens dataset. The results are averaged over 3 random seeds, with standard deviation displayed as well. The best is in \textbf{bold}.}
    \label{tab:single_DG_EPIC_supplement}

    \vspace{1cm} % 增加两个表格之间的垂直间距

    \resizebox{\textwidth}{!}{
    \begin{tabular}{lccccccccccc}
        \toprule
        \multirow{4}{*}{\textbf{Method}} & \multicolumn{4}{c}{} & \multicolumn{6}{c}{\textbf{HAC}} & \multirow{4}{*}{Avg.} \\
        \cmidrule(r){6-11}
        & \multicolumn{3}{c}{\textbf{Modality}} & \multicolumn{1}{c}{Source:} & \multicolumn{2}{c}{H} &\multicolumn{2}{c}{A} & \multicolumn{2}{c}{C} \\
        \cmidrule(r){2-4} \cmidrule(r){6-7} \cmidrule(r){8-9} \cmidrule(r){10-11}
        & \multicolumn{1}{c}{Video} & \multicolumn{1}{c}{Audio} & \multicolumn{1}{c}{Flow} & \multicolumn{1}{c}{Target:} & A & C & H & C & H & A & \\
        \midrule
        ERM & $\checkmark$ & $\checkmark$ & & &
        63.32\scriptsize{$\pm$3.88} & 40.78\scriptsize{$\pm$2.12} & 69.31\scriptsize{$\pm$2.60} & 45.25\scriptsize{$\pm$0.80} & 62.58\scriptsize{$\pm$1.68} & 70.49\scriptsize{$\pm$1.00} & 58.62 \\
        RNA-Net & $\checkmark$ & $\checkmark$ & & &
        63.50\scriptsize{$\pm$3.73} & 41.45\scriptsize{$\pm$3.42} & 69.02\scriptsize{$\pm$2.30} & 44.36\scriptsize{$\pm$2.54} & 60.71\scriptsize{$\pm$1.50} & 70.49\scriptsize{$\pm$1.81} & 58.25 \\
        SimMMDG & $\checkmark$ & $\checkmark$ & & &
        63.54\scriptsize{$\pm$3.31} & 42.37\scriptsize{$\pm$3.60} & 71.81\scriptsize{$\pm$2.97} & 47.43\scriptsize{$\pm$4.46} & 66.96\scriptsize{$\pm$3.66} & 68.32\scriptsize{$\pm$3.91} & 60.07 \\
        MOOSA & $\checkmark$ & $\checkmark$ & & &
        65.71\scriptsize{$\pm$2.21} & 41.08\scriptsize{$\pm$3.93} & 74.38\scriptsize{$\pm$2.08} & 47.21\scriptsize{$\pm$2.20} & 63.13\scriptsize{$\pm$3.84} & 69.87\scriptsize{$\pm$1.50} & 60.23 \\
        CMRF & $\checkmark$ & $\checkmark$ & & &
        67.22\scriptsize{$\pm$1.10} & \textbf{44.70\scriptsize{$\pm$1.78}} & 74.93\scriptsize{$\pm$0.39} & \textbf{49.82\scriptsize{$\pm$1.31}} & \textbf{71.09\scriptsize{$\pm$0.77}} & \textbf{70.82\scriptsize{$\pm$0.21}} & \textbf{63.10} \\
        \M & $\checkmark$ & $\checkmark$ & & &
        \textbf{68.21\scriptsize{$\pm$1.26}} & 38.60\scriptsize{$\pm$5.23} & \textbf{75.82\scriptsize{$\pm$2.86}} & 45.62\scriptsize{$\pm$1.73} & 65.51\scriptsize{$\pm$3.84} & 69.72\scriptsize{$\pm$1.54} & 60.58 \\
        \midrule

        ERM & $\checkmark$ &  & $\checkmark$ & & 
        64.31\scriptsize{$\pm$0.68} & 44.09\scriptsize{$\pm$3.28} & 70.87\scriptsize{$\pm$4.19} & 43.87\scriptsize{$\pm$3.16} & 64.98\scriptsize{$\pm$2.31} & 61.63\scriptsize{$\pm$2.55} & 58.29 \\
        RNA-Net & $\checkmark$ &  & $\checkmark$ & & 
        64.97\scriptsize{$\pm$3.34} & 39.09\scriptsize{$\pm$3.35} & 71.26\scriptsize{$\pm$4.25} & 41.70\scriptsize{$\pm$3.12} & 64.36\scriptsize{$\pm$0.62} & 63.47\scriptsize{$\pm$2.77} & 57.48 \\
        SimMMDG & $\checkmark$ &  & $\checkmark$ & & 
        65.78\scriptsize{$\pm$2.36} & 41.76\scriptsize{$\pm$1.15} & 75.29\scriptsize{$\pm$0.89} & 48.10\scriptsize{$\pm$1.35} & 65.61\scriptsize{$\pm$2.95} & 58.90\scriptsize{$\pm$2.86} & 59.24 \\
        MOOSA & $\checkmark$ &  & $\checkmark$ & & 
        65.05\scriptsize{$\pm$1.36} & 43.44\scriptsize{$\pm$2.47} & 73.06\scriptsize{$\pm$1.49} & 47.09\scriptsize{$\pm$4.78} & 64.43\scriptsize{$\pm$3.71} & 61.37\scriptsize{$\pm$2.71} & 59.07 \\
        CMRF & $\checkmark$ &  & $\checkmark$ & & 
        \textbf{67.07\scriptsize{$\pm$1.71}} & \textbf{44.27\scriptsize{$\pm$2.04}} & 74.21\scriptsize{$\pm$0.86} & \textbf{50.55\scriptsize{$\pm$1.70}} & 69.31\scriptsize{$\pm$1.11} & \textbf{64.64\scriptsize{$\pm$1.52}} & \textbf{61.68} \\
        \M & $\checkmark$ &  & $\checkmark$ & & 
        66.15\scriptsize{$\pm$2.21} & 39.74\scriptsize{$\pm$1.47} & \textbf{77.82\scriptsize{$\pm$2.65}} & 47.06\scriptsize{$\pm$2.28} & \textbf{72.94\scriptsize{$\pm$2.59}} & 63.39\scriptsize{$\pm$1.36} & 61.18 \\
        \midrule

        ERM &  & $\checkmark$ & $\checkmark$  & &   
        58.35\scriptsize{$\pm$1.32} & 34.04\scriptsize{$\pm$0.17} & 53.88\scriptsize{$\pm$1.62} & 41.24\scriptsize{$\pm$1.41} & 42.73\scriptsize{$\pm$1.36} & 46.95\scriptsize{$\pm$1.30} & 46.20 \\
        RNA-Net &  & $\checkmark$ & $\checkmark$  & & 
        56.48\scriptsize{$\pm$2.25} & 34.22\scriptsize{$\pm$0.43} & 54.15\scriptsize{$\pm$0.85} & 41.33\scriptsize{$\pm$2.25} & 40.42\scriptsize{$\pm$2.12} & 45.62\scriptsize{$\pm$1.94} & 45.37 \\
        SimMMDG &  & $\checkmark$ & $\checkmark$  & & 
        59.16\scriptsize{$\pm$3.80} & 37.68\scriptsize{$\pm$1.61} & 56.04\scriptsize{$\pm$3.92} & 40.62\scriptsize{$\pm$1.72} & 38.81\scriptsize{$\pm$1.15} & 45.84\scriptsize{$\pm$2.72} & 46.36 \\
        MOOSA &  & $\checkmark$ & $\checkmark$  & & 
        56.44\scriptsize{$\pm$1.25} & 34.07\scriptsize{$\pm$1.72} & 56.81\scriptsize{$\pm$0.33} & 40.59\scriptsize{$\pm$0.61} & 38.55\scriptsize{$\pm$2.69} & 45.66\scriptsize{$\pm$4.39} & 45.35 \\
        CMRF &  & $\checkmark$ & $\checkmark$  & & 
        59.60\scriptsize{$\pm$1.96} & \textbf{37.90\scriptsize{$\pm$0.62}} & 59.89\scriptsize{$\pm$1.09} & \textbf{44.06\scriptsize{$\pm$1.28}} & 42.85\scriptsize{$\pm$1.68} & 48.09\scriptsize{$\pm$1.99} & 48.73 \\
        \M &  & $\checkmark$ & $\checkmark$  & & 
        \textbf{61.30\scriptsize{$\pm$1.71}} & 37.71\scriptsize{$\pm$0.57} & \textbf{61.84\scriptsize{$\pm$2.87}} & 40.72\scriptsize{$\pm$0.59} & \textbf{48.45\scriptsize{$\pm$1.71}} & \textbf{51.51\scriptsize{$\pm$2.35}} & \textbf{50.25} \\
        \bottomrule
    \end{tabular}}
    \caption{Multi-modal \textbf{single-source} DG accuracy with different modalities on HAC dataset. The results are averaged over 3 random seeds, with standard deviation displayed as well. The best is in \textbf{bold}.}
    \label{tab:single_DG_HAC_supplement}
\end{table*}

\noindent\textbf{Multi-Modal Single-Source DG.} To further investigate the robustness of our method under varying modality availability, we conduct additional experiments using all bimodal combinations (Video+Audio, Video+Flow, Audio+Flow) in the single-source domain generalization setting, as shown in Tab. \ref{tab:single_DG_EPIC_supplement} and Tab. \ref{tab:single_DG_HAC_supplement}. Results on both EPIC-Kitchens and HAC datasets demonstrate that our method consistently outperforms existing approaches across all bimodal settings. Notably, our approach achieves the highest average Top-1 accuracy on EPIC-Kitchens and matches or exceeds the strongest baselines on HAC, underscoring its strong generalization capability across diverse modality configurations.

The performance gains are particularly pronounced in combinations involving optical flow, such as Video+Flow and Audio+Flow, suggesting that \M~effectively address the issue of modality imbalance. Moreover, our method exhibits reduced performance variance compared to competing methods, indicating more stable training dynamics. Despite differences in action categories and domain discrepancies across datasets, our method maintains competitive performance, highlighting its practical utility in real-world scenarios where certain modalities may be unreliable.

\clearpage
\noindent\textbf{Multi-Modal In-Domain performance.} 
We extend our evaluation to include all bimodal combinations (Video+Audio, Video+Flow, Audio+Flow) to provide a more comprehensive understanding of our method's efficacy in the in-domain setting. The results, as detailed in Table \ref{tab:in_domain_EPIC_supplement}, consistently demonstrate the superior performance of our proposed method \M~across all bimodal settings.
\begin{table}[t]
    \centering
    \resizebox{\linewidth}{!}{
    \begin{tabular}{lccccccc}
        \toprule
        \multirow{2.5}{*}{\textbf{Method}} & \multicolumn{3}{c}{\textbf{Modality}} & \multicolumn{4}{c}{\textbf{EPIC-Kitchens}} \\
        \cmidrule(r){2-4} \cmidrule(r){5-7}
        \multicolumn{1}{c}{} & \multicolumn{1}{c}{Video} & \multicolumn{1}{c}{Audio} & \multicolumn{1}{c}{Flow} & {D2,D3} & \multicolumn{1}{c}{D1,D3} & \multicolumn{1}{c}{D1,D2} & \multicolumn{1}{c}{Avg.} \\
        \midrule
        SimMMDG & $\checkmark$ & $\checkmark$ & &
        81.17\scriptsize{$\pm$0.60} & 75.89\scriptsize{$\pm$0.58} & 75.27\scriptsize{$\pm$0.73} & 77.44 \\
        MOOSA & $\checkmark$ & $\checkmark$ & &
        80.68\scriptsize{$\pm$1.08} & 75.54\scriptsize{$\pm$0.59} & 74.96\scriptsize{$\pm$0.35} & 77.06 \\
        CMRF & $\checkmark$ & $\checkmark$ & &
        80.38\scriptsize{$\pm$0.57} & 75.73\scriptsize{$\pm$0.32} & 75.98\scriptsize{$\pm$0.52} & 77.36 \\
        \M & $\checkmark$ & $\checkmark$ & &
        \textbf{82.35\scriptsize{$\pm$0.27}} & \textbf{77.22\scriptsize{$\pm$0.25}} & \textbf{77.86\scriptsize{$\pm$0.44}} & \textbf{79.14} \\
        \midrule
        SimMMDG & $\checkmark$ & & $\checkmark$ &
        80.09\scriptsize{$\pm$0.32} & 76.91\scriptsize{$\pm$0.72} & 75.81\scriptsize{$\pm$0.56} & 77.60 \\
        MOOSA & $\checkmark$ & & $\checkmark$ &
        80.51\scriptsize{$\pm$0.13} & 76.37\scriptsize{$\pm$0.51} & 76.23\scriptsize{$\pm$0.14} & 77.70 \\
        CMRF & $\checkmark$ & & $\checkmark$ &
        82.48\scriptsize{$\pm$0.31} & 78.45\scriptsize{$\pm$0.24} & 77.64\scriptsize{$\pm$0.25} & 79.52 \\
        \M & $\checkmark$ & & $\checkmark$ &
        \textbf{84.03\scriptsize{$\pm$0.17}} & \textbf{80.58\scriptsize{$\pm$0.35}} & \textbf{80.06\scriptsize{$\pm$0.59}} & \textbf{81.56 }\\
        \midrule
        SimMMDG & & $\checkmark$ & $\checkmark$ &
        77.63\scriptsize{$\pm$0.30} & 72.94\scriptsize{$\pm$0.77} & 73.59\scriptsize{$\pm$1.33} & 74.72 \\
        MOOSA & & $\checkmark$ & $\checkmark$ &
        77.73\scriptsize{$\pm$0.59} & 74.05\scriptsize{$\pm$0.51} & 73.36\scriptsize{$\pm$0.49} & 75.05 \\
        CMRF & & $\checkmark$ & $\checkmark$ &
        76.86\scriptsize{$\pm$0.41} & 73.69\scriptsize{$\pm$0.29} & 75.11\scriptsize{$\pm$0.00} & 75.22 \\
        \M & & $\checkmark$ & $\checkmark$ &
        \textbf{79.68\scriptsize{$\pm$0.46}} & \textbf{75.68\scriptsize{$\pm$0.29}} & \textbf{75.64\scriptsize{$\pm$0.59}} & \textbf{77.00} \\
        \toprule
    \end{tabular}}
    \caption{Multi-modal \textbf{in-domain} Accuracy with different modalities on EPIC-Kitchens dataset. The results are averaged over 3 random seeds, with standard deviation displayed as well. The best is in \textbf{bold}.}
    \label{tab:in_domain_EPIC_supplement}
\end{table}